\documentclass[10pt,twocolumn,letterpaper]{article}

\usepackage[T1]{fontenc}
\usepackage[pagenumbers]{wacv} % Camera-ready: authors visible, no review marks or page numbers
\usepackage{multirow}
\usepackage{tikz}
\definecolor{oursaccent}{RGB}{22,142,86}   % ViMoLa row
\definecolor{failaccent}{RGB}{200,40,40}   % artifact callouts
\DeclareMathOperator*{\argmin}{arg\,min}

\newcommand{\sys}{ViMoLa\xspace}
\newcommand{\tok}{SpatialMotionVQVAE\xspace}
\newcommand{\ttm}{TextToMotionTransformer\xspace}
\newcommand{\adapter}{WanMotionAdapter\xspace}

\usepackage{xspace}
\newcommand{\needle}{\raisebox{-0.15em}{\includegraphics[height=0.9em]{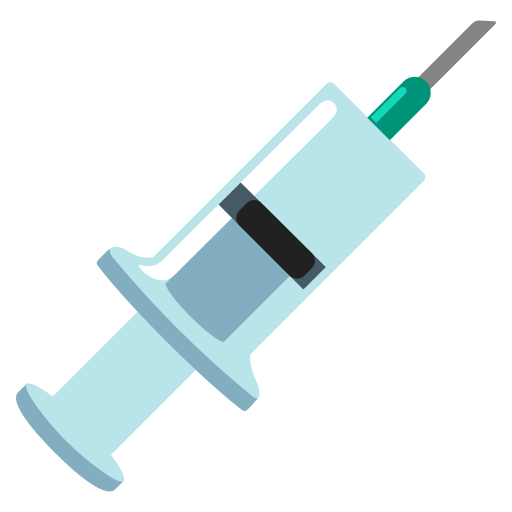}}\xspace}

\usepackage{amsmath}
\usepackage{graphicx}
\usepackage{caption}
\usepackage{xspace}
\definecolor{wacvblue}{rgb}{0.21,0.49,0.74}
\usepackage[pagebackref,breaklinks,colorlinks,allcolors=wacvblue]{hyperref}

\def\wacvPaperID{3671}
\def\confName{WACV}
\def\confYear{2027}

\newcommand{\TableMain}{%
\begin{table*}[t]
  \centering
  \caption{\textbf{Main comparison} across motion control (DA, Traj.\ L2), generative quality (FVD, Motion Smoothness), and text-video alignment (CLIP). FVD, Traj.\ L2, CLIP and smoothness use $5{,}000$ held-out clips per dataset; DA uses the fixed $300$-clip control subset, chance $=0.250$. Each method runs under its standard conditioning and none receives an explicit motion signal; DA is omitted for AnimateDiff+MotionLoRA, which takes no conditioning frame and so cannot be run on the matched first-frame subset. Best per column in \textbf{bold}.}
  \small
  \setlength{\tabcolsep}{5pt}
  \begin{tabular}{@{}llcccccc@{}}
    \toprule
    & & & & \multicolumn{2}{c}{FVD\,$\downarrow$} & & \\
    \cmidrule(lr){5-6}
    Method & Conditioning & DA\,$\uparrow$ & Traj.\ L2\,$\downarrow$ & SSV2 & K400 & CLIP\,$\uparrow$ & Motion Smooth.\,$\uparrow$ \\
    \midrule
    CogVideoX~\cite{cogvideox}                 & image\,+\,text          & $0.301$ & $0.789$ & $479.9$ & $577.1$ & $0.209$ & $0.444$ \\
    Wan~2.2 (no adapter)~\cite{wan22}           & image\,+\,text          & $0.287$ & $0.812$ & $179.4$ & \textbf{320.3} & $0.221$ & \textbf{0.472} \\
    AnimateDiff\,+\,MotionLoRA~\cite{animatediff,lora} & text only & --- & $0.820$ & $971.7$ & $454.4$ & $0.222$ & $0.367$ \\
    \midrule
    \sys (ours)                                 & image\,+\,\textbf{text} & \textbf{0.442} & \textbf{0.734} & \textbf{168.9} & $328.5$ & \textbf{0.229} & $0.455$ \\
    \bottomrule
  \end{tabular}

  \label{tab:main}
\end{table*}}

\newcommand{\TableLadder}{%
\begin{table}[t]
  \centering
  \caption{\textbf{Where the control comes from.} Same $300$ clips; chance DA $=0.250$. \textbf{Top:}~The same adapter receives across-prompt shuffled, predicted, or ground-truth codes. The oracle uses ground-truth motion at inference; other deployable rows predict motion from text. Shuffling tests prompt-matched content, not spatial arrangement. \textbf{Bottom:}~Continuous trajectory prediction from text.}
  \small
  \setlength{\tabcolsep}{3pt}
  \begin{tabular}{@{}lcc@{}}
    \toprule
    Motion source & DA\,$\uparrow$ & Traj.\ L2\,$\downarrow$ \\
    \midrule
    no adapter               & $0.287$ & $0.812$ \\
    shuffled codes           & $0.283$ & $0.806$ \\
    \textbf{predicted codes} & $\mathbf{0.442}$ & $\mathbf{0.734}$ \\
    ground truth (oracle)    & $0.531$ & $0.702$ \\
    \midrule
    cont.\ T2M\,$\to$\,SG-I2V~\cite{sgi2v} & $0.352$ & $0.781$ \\
    cont.\ T2M\,$\to$\,our adapter         & $0.331$ & $0.792$ \\
    \bottomrule
  \end{tabular}

  \label{tab:ladder}
\end{table}}

\newcommand{\TableCost}{%
\begin{table}[t]
  \centering
  \caption{\textbf{Cost of control.} Matched training budget; FLOPs are relative to a frozen DiT forward pass. Inference overhead is measured per clip on one H100. Control is the DA gap over across-prompt shuffled codes (\ifdefined\StandaloneSupplement main-paper Table~2\else\cref{tab:ladder}\fi).}
  \footnotesize
  \setlength{\tabcolsep}{3pt}
  \begin{tabular}{@{}lccccc@{}}
    \toprule
    Variant & Params & \% FLOPs & Infer.\ $\Delta$ & VRAM & DA gap \\
    \midrule
    Full                & $1.14$B & $7.88\%$ & $+0.31$\,s & $62.4$\,GB & $+0.159$ \\
    Rank $256$ & $\mathbf{201}$\textbf{M} & $\mathbf{1.45\%}$ & $\mathbf{+0.09}$\,\textbf{s} & $\mathbf{41.7}$\,\textbf{GB} & $+0.145$ \\
    \bottomrule
  \end{tabular}
  \label{tab:cost}
\end{table}}

\newcommand{\TableComplex}{%
\begin{table}[t]
  \centering
  \caption{\textbf{Multi-phase and compound motion} at $s{=}0.06$; chance DA $=0.250$. Top: separate split sizes and end-to-end DA gains. Bottom: tokenizer reconstruction on the multi-phase split, using $S$ temporal segments. Segmentation improves reconstruction; generation with segmented tokens was not evaluated.}
  \small
  \setlength{\tabcolsep}{4pt}
  \begin{tabular}{@{}lcccc@{}}
    \toprule
    Split & $n$ & \sys\,DA\,$\uparrow$ & Wan\,DA\,$\uparrow$ & $\Delta$ \\
    \midrule
    single-phase       & $300$ & $\mathbf{0.478}$ & $0.291$ & \textbf{+0.187} \\
    multi-phase (auto) & $252$ & $0.334$ & $0.286$ & $+0.048$ \\
    compound (curated) & $150$ & $0.307$ & $0.253$ & $+0.054$ \\
    \bottomrule
  \end{tabular}

  \vspace{6pt}
  \begin{tabular}{@{}lc@{}}
    \toprule
    Tokenization (multi-phase split) & Recon.\ EPE\,$\downarrow$ \\
    \midrule
    $64$ spatial codes (ours)                      & $0.147$ \\
    $64{\times}2$ spatiotemporal                   & $0.104$ \\
    $\mathbf{64{\times}4}$ \textbf{spatiotemporal} & $\mathbf{0.089}$ \\
    \bottomrule
  \end{tabular}

  \label{tab:complex}
\end{table}}

\newcommand{\TableMotion}{%
\begin{table}[t]
  \centering
  \caption{\textbf{Text-to-motion by caption type}, $K{=}512$. SSV2's templated captions induce a \emph{higher-entropy}, harder code distribution than K400's, with a far weaker majority-code prior, which is why SSV2's raw Top-1 is lower despite its cleaner supervision. Measured as lift over that prior, the ordering inverts and matches intuition. Pooled over both splits, Top-1 is $8.6\%$ and Top-5 $16.6\%$. The codebook-size sweep behind these numbers is in \cref{tab:codebook}.}
  \footnotesize
  \setlength{\tabcolsep}{3pt}
  \begin{tabular}{@{}lccccc@{}}
    \toprule
    Split & Top-1\,$\uparrow$ & Top-5\,$\uparrow$ & $H$ & Prior & Lift\,$\uparrow$ \\
    \midrule
    SSV2 (templated) & $7.5\%$           & $15.4\%$          & $5.77$ & $2.4\%$  & $\mathbf{3.1\times}$ \\
    K400 (free-form) & $\mathbf{15.9\%}$ & $\mathbf{24.5\%}$ & $5.37$ & $12.7\%$ & $1.25\times$ \\
    \bottomrule
  \end{tabular}

  \label{tab:t2m}
\end{table}}

\newcommand{\TableQuality}{%
\begin{table}[t]
  \centering
  \caption{\textbf{Quality and identity} at $s{=}0.06$. Brackets are supplied $95\%$ confidence intervals, which overlap for SSV2 FVD and DINO subject consistency. Other quality scores are point estimates. DA is reported over three seeds; frame-$0$ error measures deviation from the conditioning image.}
  \small
  \setlength{\tabcolsep}{4pt}
  \begin{tabular}{@{}lcc@{}}
    \toprule
    Metric & \sys & Wan~2.2 \\
    \midrule
    FVD SSV2\,$\downarrow$        & $\mathbf{168.9}$ {\scriptsize$[161,177]$} & $179.4$ {\scriptsize$[172,188]$} \\
    FVD K400\,$\downarrow$        & $328.5$ & $\mathbf{320.3}$ \\
    CLIP\,$\uparrow$              & $\mathbf{0.229}$ & $0.221$ \\
    Motion Smooth.\,$\uparrow$    & $0.455$ & $\mathbf{0.472}$ \\
    DINO subj.\ cons.\,$\uparrow$ & $0.731$ {\scriptsize$[.712,.749]$} & $\mathbf{0.743}$ {\scriptsize$[.725,.760]$} \\
    Backgr.\ cons.\,$\uparrow$    & $0.812$ & $\mathbf{0.821}$ \\
    Frame-$0$ $L_2$               & ${\sim}10^{-4}$ & ${\sim}10^{-4}$ \\
    \midrule
    DA, $3$ seeds\,$\uparrow$     & $\mathbf{0.442 \pm 0.011}$ & $0.287 \pm 0.009$ \\
    \bottomrule
  \end{tabular}

  \label{tab:quality}
\end{table}}

\newcommand{\TableAblation}{%
\begin{table}[t]
  \centering
  \caption{\textbf{Adapter design and operating point}, $s{=}0.06$ unless swept; chance DA $=0.250$. \textbf{(top)}~Ablations over the choices of \cref{sec:adapter}; the first row of each group repeats the default of \cref{tab:main}. $^{\dagger}$Inference-time probe. \textbf{(bottom)}~The knob $s$ rescales every block's gate, so control and quality read off a single row.}
  \small
  \setlength{\tabcolsep}{3pt}
  \begin{tabular}{@{}llccc@{}}
    \toprule
    & Variant & FVD\,$\downarrow$ & Motion\,$\uparrow$ & CLIP\,$\uparrow$ \\
    \midrule
    \multirow{2}{*}{Gate}
      & zero-gated (ours) & \textbf{168.9} & $0.455$ & \textbf{0.229} \\
      & ungated$^\dagger$ & $7576.1$ & $0.467$ & $0.155$ \\
    \midrule
    \multirow{3}{*}{Injection}
      & all $30$ blocks (ours)      & \textbf{168.9} & $0.455$ & \textbf{0.229} \\
      & first $15$ blocks$^\dagger$ & $542.1$ & $0.464$ & $0.219$ \\
      & last $15$ blocks$^\dagger$  & $511.6$ & \textbf{0.488} & $0.224$ \\
    \bottomrule
  \end{tabular}

  \vspace{6pt}
  \begin{tabular}{@{}lccccc@{}}
    \toprule
    $s$ & DA\,$\uparrow$ & CLIP\,$\uparrow$ & Smooth.\,$\uparrow$ & FVD\,$\downarrow$ & DINO\,$\uparrow$ \\
    \midrule
    $0.01$ & $0.312$ & \textbf{0.232} & \textbf{0.469} & $174.1$ & \textbf{0.741} \\
    $0.03$ & $0.389$ & $0.231$ & $0.464$ & $171.6$ & $0.737$ \\
    $\mathbf{0.06}$ \textbf{(def.)} & $0.442$ & $0.229$ & $0.455$ & $\mathbf{168.9}$ & $0.731$ \\
    $0.10$ & \textbf{0.451} & $0.216$ & $0.398$ & $189.4$ & $0.702$ \\
    \bottomrule
  \end{tabular}

  \label{tab:ablation}
\end{table}}

\newcommand{\TableParams}{%
\begin{table}[h]
  \centering
  \caption{\textbf{Parameter budget.} Only the per-block adapters are trained during the final stage; all backbone weights remain frozen. $^{\dagger}$Trained in an earlier stage and frozen here. The low-rank adapter ($r{=}256$, main-paper Sec.~3.4) is a drop-in replacement for the full adapter at $18\%$ of its trainable size; \cref{tab:cost} reports the compute cost of both.}
  \small
  \setlength{\tabcolsep}{6pt}
  \begin{tabular}{@{}lcc@{}}
    \toprule
    Component & Status & Params \\
    \midrule
    \tok                              & Frozen$^{\dagger}$        & $2.7$M \\
    \ttm                              & Frozen$^{\dagger}$        & $27.6$M \\
    \adapter ($\times 30$)            & \textbf{Trained}          & \textbf{1{,}144.6M} \\
    \quad low-rank, $r{=}256$         & \textbf{Trained}          & \textbf{201.0M} \\
    \midrule
    Wan~2.2 DiT                       & Frozen                    & $5{,}000$M \\
    Wan~2.2 VAE                       & Frozen                    & $704.7$M \\
    UMT5-XXL text encoder             & Frozen                    & $5{,}680.9$M \\
    \midrule
    \textbf{Trainable (full adapter)} &                           & \textbf{1{,}144.6M} ($9.1\%$) \\
    \textbf{Trainable (low-rank)}     &                           & \textbf{201.0M} ($1.6\%$) \\
    \bottomrule
  \end{tabular}
  \label{tab:params}
\end{table}}

\newcommand{\TableCodebook}{%
\begin{table}[h]
  \centering
  \caption{\textbf{Why a low Top-1 is not a low-fidelity motion prior.} Codebook size $K$ at $d{=}128$, on $4{,}000$ validation clips. Hard Top-1 falls by $4.3\times$ as the vocabulary grows, while every measure of the \emph{decoded} motion --- end-point error, directional accuracy, and the tolerance-based $\varepsilon$-soft Top-1 (\cref{sec:supp_protocol}) --- stays flat: at large $K$ the codebook holds many near-duplicate patterns, so an ``incorrect'' code is usually a near-synonym of the correct one. We adopt $K{=}512$, where reconstruction has largely converged without the diminishing returns of $K{=}1024$.}
  \small
  \setlength{\tabcolsep}{5pt}
  \begin{tabular}{@{}lccccc@{}}
    \toprule
    $K$ & Recon.\ EPE\,$\downarrow$ & Top-1\,$\uparrow$ & $\varepsilon$-soft Top-1\,$\uparrow$ & Decoded EPE\,$\downarrow$ & DA\,$\uparrow$ \\
    \midrule
    $128$          & $0.112$          & $\mathbf{22.1\%}$ & $\mathbf{0.594}$ & $0.198$          & $0.431$ \\
    $256$          & $0.096$          & $14.2\%$          & $0.548$          & $0.191$          & $0.438$ \\
    $\mathbf{512}$ & $\mathbf{0.084}$ & $8.6\%$           & $0.512$          & $0.188$          & $0.442$ \\
    $1024$         & $0.079$          & $5.1\%$           & $0.487$          & $\mathbf{0.187}$ & $\mathbf{0.445}$ \\
    \bottomrule
  \end{tabular}
  \label{tab:codebook}
\end{table}}

\newcommand{\TableHuman}{%
\begin{table}[h]
  \centering
  \caption{\textbf{Blind pairwise human study.} $7$ raters $\times$ $150$ pairs, matched seeds and matched first frames, so the two clips in a pair differ only in whether the adapter was attached. Raters prefer \sys on direction and on speed and are at chance on overall quality --- the same conclusion the automatic metrics of \cref{tab:quality} reach, from an independent source.}
  \small
  \setlength{\tabcolsep}{6pt}
  \begin{tabular}{@{}lcc@{}}
    \toprule
    Question & \sys win rate & $95\%$ CI \\
    \midrule
    Direction match & $\mathbf{68.4\%}$ & $[63.1,\,73.3]$ \\
    Speed match     & $\mathbf{61.2\%}$ & $[55.7,\,66.4]$ \\
    Overall quality & $49.1\%$          & $[43.6,\,54.6]$ \\
    \bottomrule
  \end{tabular}
  \label{tab:human}
\end{table}}

\newcommand{\TableTokenizerDim}{%
\begin{table}[h]
  \centering
  \caption{\textbf{Motion tokenizer, latent dimension.} Reconstruction error and codebook utilization at $K{=}512$ as the latent width $d$ varies, on the full held-out set. $d{=}128$ gives the best reconstruction $L_2$ and by far the healthiest codebook occupancy, which is why we adopt it; the codebook-size sweep at $d{=}128$ is in \cref{tab:codebook}.}
  \small
  \setlength{\tabcolsep}{5pt}
  \begin{tabular}{@{}ccccc@{}}
    \toprule
    Codebook $K$ & Latent $d$ & \# Tokens & Recon.\ $L_2$\,$\downarrow$ & Perplexity\,$\uparrow$ \\
    \midrule
    $512$          & $64$           & $64$          & $3.19{\times}10^{-4}$          & $168.2$ \\
    $\mathbf{512}$ & $\mathbf{128}$ & $\mathbf{64}$ & $\mathbf{3.15{\times}10^{-4}}$ & $\mathbf{290.7}$ \\
    $512$          & $256$          & $64$          & $3.19{\times}10^{-4}$          & $171.2$ \\
    \bottomrule
  \end{tabular}
  \label{tab:tokenizerdim}
\end{table}}

\newcommand{\TablePriors}{%
\begin{table}[h]
  \centering
  \caption{\textbf{What the predicted codes are worth in motion space.} End-point error of the field decoded from each source of codes, $K{=}512$, $4{,}000$ validation clips. The predicted codes recover most of the distance between an uninformative prior and the tokenizer's own ceiling, even though their hard Top-1 is $8.6\%$.}
  \small
  \setlength{\tabcolsep}{6pt}
  \begin{tabular}{@{}lc@{}}
    \toprule
    Conditioning & Decoded EPE\,$\downarrow$ \\
    \midrule
    random code                           & $0.417$ \\
    per-patch majority code               & $0.291$ \\
    \textbf{predicted code (\ttm)}        & $\mathbf{0.188}$ \\
    ground-truth code (tokenizer ceiling) & $0.084$ \\
    \bottomrule
  \end{tabular}
  \label{tab:priors}
\end{table}}

\newcommand{\TableBackbone}{%
\begin{table}[t]
  \centering
  \caption{\textbf{Transfer to CogVideoX-5B-I2V.} The tokenizer and text-to-motion predictor are reused unchanged; only generator-side adapters are retrained. DA improves on both backbones, with a smaller gain on CogVideoX. DINO SC is subject consistency; chance DA $=0.250$.}
  \footnotesize
  \setlength{\tabcolsep}{3pt}
  \begin{tabular}{@{}llccc@{}}
    \toprule
    Backbone & Adapter & DA\,$\uparrow$ & $\Delta$DA & DINO SC\,$\uparrow$ \\
    \midrule
    \multirow{2}{*}{Wan~2.2 5B} & none & $0.287$ & --- & $0.743$ \\
                                     & \sys & $0.442$ & \textbf{+0.155} & $0.731$ \\
    \midrule
    \multirow{2}{*}{CogVideoX 5B} & none & $0.301$ & --- & $0.795$ \\
                                      & \sys & $0.398$ & \textbf{+0.097} & $0.788$ \\
    \bottomrule
  \end{tabular}
  \label{tab:backbone}
\end{table}}

\newcommand{\TableComplexMain}{%
\begin{table}[t]
\centering
\caption{\textbf{Complex-motion control.} DA at $s{=}0.06$, chance $0.250$. Split construction and tokenizer reconstruction diagnostics are in \cref{tab:complex}.}
\small
\setlength{\tabcolsep}{5pt}
\begin{tabular}{@{}lrccc@{}}
\toprule
Split & $n$ & Wan & \sys & $\Delta$ \\
\midrule
Single-phase & $300$ & $.291$ & $.478$ & $+.187$ \\
Multi-phase & $252$ & $.286$ & $.334$ & $+.048$ \\
Compound & $150$ & $.253$ & $.307$ & $+.054$ \\
\bottomrule
\end{tabular}
\label{tab:complexmain}
\end{table}}

\newcommand{\TableCodebookMain}{%
\begin{table}[t]
\centering
\caption{\textbf{Codebook size and motion fidelity.} Hard token accuracy falls as $K$ grows, while decoded motion EPE and DA remain stable (chance DA $=0.250$). Token diagnostics use $4{,}000$ validation clips; bold marks the default $K{=}512$. Reconstruction and $\varepsilon$-soft accuracy are reported in \cref{tab:codebook}.}
\small
\setlength{\tabcolsep}{5pt}
\begin{tabular}{@{}rccc@{}}
\toprule
$K$ & Token Top-1$\uparrow$ & Decoded EPE$\downarrow$ & DA$\uparrow$ \\
\midrule
$128$ & $22.1\%$ & $0.198$ & $0.431$ \\
$256$ & $14.2\%$ & $0.191$ & $0.438$ \\
$\mathbf{512}$ & $\mathbf{8.6\%}$ & $\mathbf{0.188}$ & $\mathbf{0.442}$ \\
$1024$ & $5.1\%$ & $0.187$ & $0.445$ \\
\bottomrule
\end{tabular}
\label{tab:codebookmain}
\end{table}}

\newcommand{\FigPipeline}{
\begin{figure*}[t]
  \centering
  \includegraphics[width=0.94\linewidth]{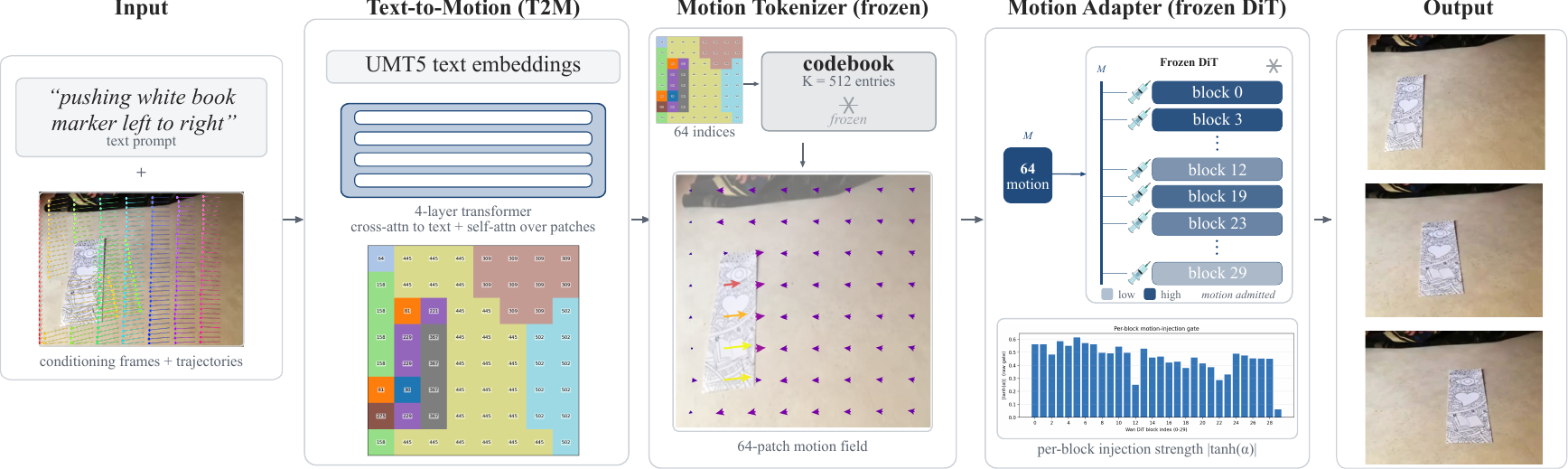}
  \caption{\textbf{\sys overview.} At inference, a text prompt is encoded by UMT5 and mapped by \ttm to
  $64$ discrete motion-token indices over a $512$-entry codebook
  (\emph{Text-to-Motion}). A frozen codebook lookup (\tok) turns these into the
  $64$-patch motion field they encode (\emph{Motion Tokenizer}). A zero-gated
  cross-attention \adapter then \needle injects the motion embeddings into every
  block of the frozen Wan~2.2 DiT through forward hooks (\emph{Motion
  Adapter}); block shading and the bar chart show the learned per-block
  injection strength $|\tanh(\alpha)|$, which varies across depth. Note the
  motion field: marker patches point right as prompted while the surrounding
  scene points left. The decoded grid illustrates the tokenizer representation; the adapter receives its embeddings without explicit patch positions. The frozen DiT then denoises and its VAE decodes the clip
  (\emph{Output}), here ``pushing white book marker left to right'' on a
  held-out SSV2 example. The first frame conditions the backbone, not \ttm. Only the adapters are trained in the final stage.}
  \label{fig:pipeline}
\end{figure*}
}

\newcommand{\ViMolaQual}{
\begin{figure*}[t!] % Added ! for better placement flexibility
  \centering
  \includegraphics[width=0.70\textwidth]{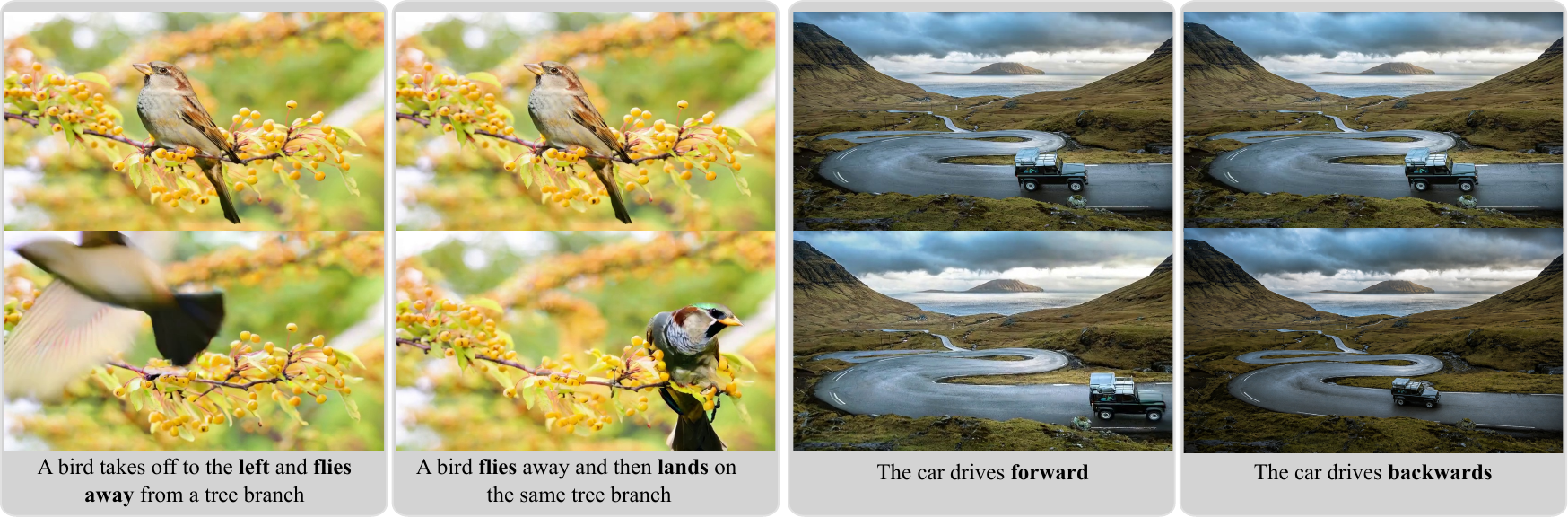}
  \caption{\textbf{\sys qualitative results} on in-the-wild images sourced from
  the internet, not from SSV2 or K400. Across ``a bird takes off to the left''
  \vs ``a bird flies away and then lands on the same tree branch,'' and ``the car
  drives forward'' \vs ``backwards'' on the same road, the prompts vary the requested motion within a scene. These examples illustrate behavior; average subject consistency is measured in \cref{tab:quality} and does not establish object binding. Failure cases, concentrated on the compound prompts of
  \cref{tab:complex}, and the learned per-block gates are in
  \cref{sec:supp_failures}.}
  \label{fig:qualitative}
\end{figure*}
}

\title{ViMoLa: Discrete Motion Tokens for Language-Guided Video Generation}

\author{%
  Bishoy Galoaa\\
  Northeastern University\\
  {\tt\small galoaa.b@northeastern.edu}
  \and
  Sarah Ostadabbas\thanks{Corresponding author.}\\
  Northeastern University\\
  {\tt\small s.ostadabbas@northeastern.edu}
}

\begin{document}
\twocolumn[{
\renewcommand\twocolumn[1][]{#1}
\maketitle
\begin{center}
    \vspace{-28pt}
    \includegraphics[width=1.0\linewidth]{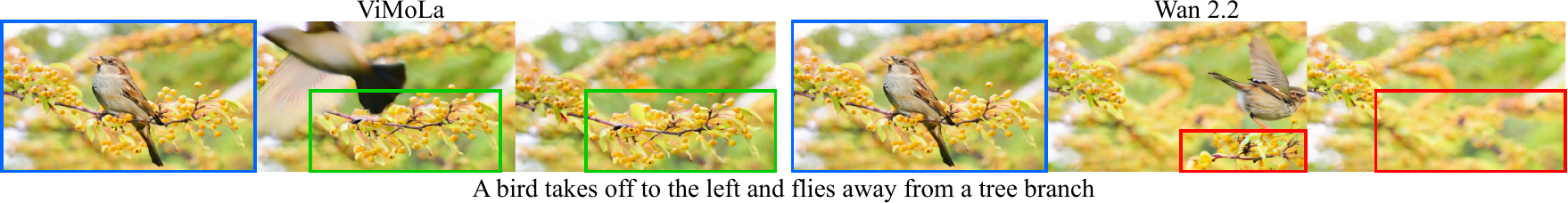}
    \centering
    \captionsetup{type=figure}
    \vspace{-5pt}
    \caption{\textbf{Language-guided motion.} For ``a bird takes off to the left and flies away from a tree branch,'' \sys (left) follows the prompted direction, while the frozen Wan~2.2 baseline (right) moves right and deforms the branch. Both receive the same image and text; \sys adds a text-predicted motion prior. This example illustrates directional control, rather than establishing object-specific grounding.}
    \label{fig:teaser}
\end{center}
}]

\begin{abstract}
Language-guided video generation often captures scene content while missing the requested motion. We present \sys (Video Motion Language), which predicts a motion prior from language and injects it into a frozen image-to-video backbone. A spatial vector-quantized autoencoder compresses dense trajectories into $64$ tokens from a $512$-entry codebook; a $28$M-parameter transformer predicts the tokens from text; and zero-gated cross-attention adapters condition every frozen Wan~2.2 DiT block through forward hooks. Trained on $462$K videos from Something-Something~v2 and Kinetics-400, \sys raises directional accuracy from $0.287$ to $0.442$ (chance $0.250$), with similar measured generative quality. Across-prompt shuffled codes achieve $0.283$, supporting the importance of prompt-matched token content. A low-rank adapter reaches $0.428$ at $1.45\%$ additional forward-pass FLOPs, and reusing the tokenizer and language model with newly trained CogVideoX adapters raises accuracy from $0.301$ to $0.398$. Control remains imperfect: oracle codes reach $0.531$, compound-motion gains are smaller, and the adapter does not explicitly preserve token-to-patch assignments.
Code is available at \href{https://github.com/ostadabbas/ViMoLa}{\texttt{github.com/ostadabbas/ViMoLa}}.
\end{abstract}

\section{Introduction}
\label{sec:intro}

Text-to-video generation has advanced rapidly, with large diffusion
transformers (DiTs) such as Wan~2.2~\cite{wan2025,wan22}, CogVideoX~\cite{cogvideox},
and Stable Video Diffusion~\cite{svd} producing clips that are sharp,
temporally coherent, and rich in motion. These models are exceptionally good at
deciding \emph{what} a video contains: prompt them for ``a hand sliding a cup
across a table'' and a hand and a cup reliably appear. They are far less
controllable about \emph{how} the content moves. The same prompt may send the
cup left, right, or barely at all; the trajectory, speed, and timing of the
motion are an emergent property of sampling rather than something the user can
specify (\cref{fig:teaser}). For applications that care about motion (e.g., robotics demonstrations,
instructional content, animation, data augmentation for action
recognition) this gap between semantic and \emph{kinematic} control is a fundamental limitation.

A growing body of work restores motion control by conditioning generation on
an explicit motion signal: trajectories, drags, or dense flow fields fed
directly into the denoiser~\cite{motionctrl,draganything,dragnuwa,tora,motionprompting}.
These methods are powerful, but they share two properties that limit their
reach. First, the user must supply that motion at inference time. Dense
motion is exactly the kind of high-dimensional, hard-to-author signal most
users do not have, which is why language became the dominant interface for
generation in the first place. Second, control is typically wired into the
base generator, either by adding trainable branches to its layers or by
fine-tuning it, which ties the controller to a specific backbone and risks
eroding the generative prior that makes the backbone valuable.

We learn a compact motion vocabulary that bridges language and a frozen image-to-video generator. Neighboring tracked points often move coherently, allowing a spatial vector-quantized autoencoder to compress a dense trajectory field into $64$ patch codes from a $512$-entry vocabulary. A $28$M transformer predicts these codes from UMT5~\cite{umt5} text embeddings. Zero-gated cross-attention adapters then condition the frozen Wan~2.2 DiT through forward hooks, inspired by zero-initialized control pathways~\cite{controlnet,t2iadapter}. Removing the adapters exactly recovers the original model.

Our contributions are:
\begin{itemize}
\item \textbf{A language-predicted discrete motion prior.} Motion-code classification enables directional and speed control without user-supplied tracks, drags, or flow. The predictor uses text and a fixed patch grid, rather than image features; object-specific binding is not established by our evaluation.
\item \textbf{Reversible conditioning of frozen video backbones.} A low-rank adapter trains $201$M parameters and adds $1.45\%$ to forward-pass FLOPs, with DA $0.428$ versus $0.442$ for the full adapter. The same tokenizer and predictor transfer to CogVideoX when only adapters are retrained.
\item \textbf{Controlled evaluation of gains and limits.} Predicted, across-prompt shuffled, and oracle codes separate the effect of prompt-matched motion content from an uninformative conditioning signal. Continuous trajectory baselines, human judgments, and compound-motion diagnostics complement the main comparison. These tests support directional control but do not isolate spatial grounding or the full source of the oracle gap.
\end{itemize}

\section{Related Work}
\label{sec:related}

\subsection{Text- and image-to-video generation}
Video diffusion models have scaled from U-Net latent-diffusion
designs~\cite{ldm,svd} to diffusion transformers (DiT)~\cite{dit} trained at
scale. Recent open systems such as CogVideoX~\cite{cogvideox} and
Wan~2.2~\cite{wan2025,wan22} produce high-resolution, temporally consistent
clips and are increasingly trained with flow-matching objectives~\cite{flowmatching,rectifiedflow}.
Wan~2.2 in particular is a strong, openly available backbone: a $3$D DiT with a
spatiotemporal VAE and a UMT5~\cite{umt5} text encoder, usable in text-to-video
or image-to-video mode. These models excel at semantic faithfulness but expose
only the prompt, and optionally a starting frame, as control surfaces, neither of
which determines scene motion. \sys treats one such model, in image-to-video
mode, as a \emph{frozen} backbone and adds motion controllability around it.

\subsection{Motion- and trajectory-controlled generation}
A large body of work makes video generation controllable by conditioning on an
explicit motion signal. MotionCtrl~\cite{motionctrl} disentangles camera and
object motion and injects both into a latent video diffusion model;
DragNUWA~\cite{dragnuwa} and DragAnything~\cite{draganything} convert
user-drawn drags or entity trajectories into spatial conditioning;
SparseCtrl~\cite{sparsectrl} and CameraCtrl~\cite{cameractrl} add sparse
structural or camera controls; and Trajectory Attention~\cite{trajattn},
SG-I2V~\cite{sgi2v}, and MagicMotion~\cite{magicmotion} steer motion through
attention or dense-to-sparse trajectory guidance. Closest in spirit,
Tora~\cite{tora} encodes trajectories into spacetime motion patches and fuses
them into a DiT, and Motion Prompting~\cite{motionprompting} conditions on dense
point tracks as a general motion interface.

Two distinctions separate \sys from this family. \emph{(i)~Conditioning
modality.} All of the above require an explicit motion input at inference,
such as a trajectory, a drag, or a dense track field, whereas \sys requires
no such motion signal and \emph{predicts} the motion from a text prompt via a
dedicated text-to-motion stage. This
makes the systems complementary: a method like Tora or Motion Prompting could in
principle consume the tracks our tokenizer decodes, but neither offers a
language-to-motion pathway. \emph{(ii)~Backbone treatment.} Many controllers add
trainable branches to, or fine-tune, the base U-Net or DiT. \sys keeps the
backbone entirely frozen and attaches non-invasive adapters via forward hooks,
preserving the pretrained prior and remaining portable across backbones. Since
these methods require ground-truth motion at inference, the fair text-only
comparison is a text-to-trajectory front-end feeding a trajectory-conditioned
generator, which we build and evaluate in \cref{sec:experiments}.

\subsection{Parameter-efficient adaptation of diffusion models}
Adapters add new capabilities to a frozen generator at a fraction of the cost of
full fine-tuning. ControlNet~\cite{controlnet} clones encoder blocks and gates
them with zero-initialized convolutions so that training begins from the
unmodified model; T2I-Adapter~\cite{t2iadapter} learns lightweight structural
adapters; and IP-Adapter~\cite{ipadapter} introduces decoupled cross-attention
for image prompts. In the video setting, AnimateDiff~\cite{animatediff} inserts
trainable temporal motion modules into a frozen image model and adapts to
specific motions with LoRA~\cite{lora}. \sys adopts the zero-gated philosophy of
ControlNet but differs in mechanism and modality: it adds a \emph{parallel
cross-attention} pathway, rather than a cloned branch or residual convolution,
whose key/value memory is a sequence of discrete \emph{motion} tokens, and it
attaches through forward hooks so the base code is never modified.

\subsection{Discrete representation learning}
Vector quantization~\cite{vqvae} learns a discrete codebook bottleneck and has
become a standard tool for compact, generative-friendly representations;
VQGAN~\cite{vqgan} extends it to spatially discrete image tokens that downstream
models can predict autoregressively. \sys applies this idea to a new modality:
dense spatiotemporal \emph{motion} fields. Rather than tokenizing pixels, \tok
tokenizes per-patch velocity, summarizing each patch's motion over the clip as a
single code. The resulting vocabulary is what lets a small transformer predict
motion from text by classification, mirroring how discrete image tokens enable
language-style modeling of appearance.

\section{Method}
\label{sec:method}

\FigPipeline

\subsection{Overview}

\sys factorizes language-guided motion control into three components trained in
sequence on top of a frozen Wan~2.2 backbone (\cref{fig:pipeline}). A spatial
motion tokenizer (\tok, \cref{sec:tokenizer}) learns a discrete vocabulary of
motion by compressing dense point-trajectory fields into $64$ codebook tokens. A
\ttm (\cref{sec:t2m}) learns to predict those tokens from natural language.
A \adapter (\cref{sec:adapter}) injects the resulting motion prior into every
block of the frozen DiT. Training proceeds in four stages: \tok on trajectory
data; cached UMT5 text embeddings; \ttm against frozen tokenizer targets; and
the adapters under a flow-matching objective on pre-encoded VAE latents. Stages
decouple cleanly, so each component can be developed and debugged
independently.

At inference, motion is driven by text alone: no trajectory, drag, or flow field
is required. A prompt is encoded by UMT5; \ttm predicts $64$ motion-token
indices; a codebook lookup turns these into motion embeddings; the \adapter
conditions each DiT block on them; and Wan~2.2, run image-to-video from a single
conditioning frame, denoises a latent its VAE decodes into frames.

\subsection{Spatial motion tokenizer}
\label{sec:tokenizer}

\paragraph{Motion representation.}
Given a clip of $T{+}1$ frames, we track $N{=}1024$ points arranged on a
$32{\times}32$ grid with CoTracker3~\cite{cotracker3} and represent motion as
frame-to-frame velocities,
\begin{equation}
  \mathbf{V} \in \mathbb{R}^{T \times N \times 2},
  \qquad
  \mathbf{p}_0 \in \mathbb{R}^{N \times 2},
  \label{eq:motionfield}
\end{equation}
where $T{=}48$ (from $49$ frames) and $\mathbf{p}_0$ are the initial point
positions, normalized to $[-1,1]^2$. Velocities rather than absolute positions
make the representation translation-invariant and zero-centered, which suits
quantization.

\paragraph{Spatial patch decomposition.}
We partition the grid into $P{=}64$ non-overlapping $4{\times}4$ patches of $16$
points each, so each patch carries a local velocity field
$\mathbf{V}^{(j)} \in \mathbb{R}^{T \times 16 \times 2}$. This encodes our locality assumption into the architecture: because neighboring
points move coherently, a patch's motion is well described by a single dominant
pattern, and patches can be encoded independently with a shared network. The temporal dimension is summarized within each patch, so the $64$
tokenizer output retains patch assignments for reconstructing the field, rather than
explicitly allocating separate codes to successive motion phases, a deliberate trade that favors scenes with coherent
regional motion. \cref{sec:complex} measures what this costs on multi-phase
motion and shows that the same tokenizer, given a temporal axis of $S$ sub-clip
segments, recovers the lost fidelity at $64{\times}S$ codes; we quantify the
$S{=}1$ configuration end to end here (\cref{sec:limitations}). Because patches are independent,
foreground and background can carry opposite codes without conflict, as in
\cref{fig:pipeline}.

\paragraph{Encoder, quantization, and decoder.}
A shared patch encoder $E$ maps each patch to a latent
$\mathbf{z}^{(j)} \in \mathbb{R}^{d}$ with $d{=}128$: it flattens the spatial
dimension, applies a small temporal $1$D CNN with GELU activations, pools over
time, and projects with an MLP, with all $P$ patches processed in parallel. Each
latent is quantized to its nearest codebook entry,
\begin{equation}
  \mathbf{z}_q^{(j)} = \mathbf{e}_{k^\star}, \qquad
  k^\star = \argmin_{k \in \{1,\dots,K\}} \big\lVert \mathbf{z}^{(j)} - \mathbf{e}_k \big\rVert_2,
  \label{eq:vq}
\end{equation}
with a codebook of $K{=}512$ entries $\{\mathbf{e}_k\}\subset\mathbb{R}^{d}$. The
codebook is updated with exponential moving averages (decay $0.99$) and
dead-code reset, and trained with the standard commitment objective. A symmetric
patch decoder $D$ reconstructs the per-patch velocity field, and integrating the
predicted velocities from $\mathbf{p}_0$ recovers full trajectories. The
tokenizer is trained by
\begin{equation}
  \mathcal{L}_{\text{tok}} =
  \underbrace{\big\lVert \mathbf{V} - \hat{\mathbf{V}} \big\rVert_2^2}_{\text{reconstruction}}
  + \underbrace{\big\lVert \operatorname{sg}[\mathbf{z}] - \mathbf{z}_q \big\rVert_2^2}_{\text{codebook}}
  + \beta \underbrace{\big\lVert \mathbf{z} - \operatorname{sg}[\mathbf{z}_q] \big\rVert_2^2}_{\text{commitment}},
  \label{eq:tokloss}
\end{equation}
with $\beta{=}0.25$ and $\operatorname{sg}[\cdot]$ the stop-gradient. The encoder
emits per-clip indices $\mathbf{c}\in\{1,\dots,K\}^{P}$ and quantized embeddings
$\mathbf{Z}_q\in\mathbb{R}^{P\times d}$; the former serve as classification
targets for \ttm, the latter as the conditioning memory for the adapter.

\subsection{Text-to-motion transformer}
\label{sec:t2m}
\ttm predicts the $64$-token motion code from a caption and the initial point
layout. It receives UMT5 per-token text embeddings
$\mathbf{H}\in\mathbb{R}^{L\times 4096}$ ($L\le 77$) and the initial positions,
pooled from $N{=}1024$ points to $P{=}64$ patch centers. Each patch center is
given a $2$D sinusoidal positional encoding and a learned embedding, forming
$64$ patch queries in a model of width $512$. Four transformer layers then
alternate (a)~cross-attention from the patch queries into the text embeddings,
letting language steer the motion, and (b)~self-attention among the patch
queries, enforcing spatial coherence, each followed by a feed-forward block. A
linear head produces per-patch logits over the $K{=}512$ codes. The layout is a fixed grid; the predictor does not receive conditioning-frame features, so its patch queries cannot directly bind a named object to its location in a particular image.

Because the tokenizer makes the target discrete, training is simply
cross-entropy against the frozen tokenizer indices,
\begin{equation}
  \mathcal{L}_{\text{T2M}}
  = \frac{1}{P}\sum_{j=1}^{P}
  \operatorname{CE}\!\big(\,\boldsymbol{\ell}^{(j)},\; c^{(j)}\,\big),
  \qquad
  \mathbf{c} = E_{\text{VQ}}(\mathbf{V}),
  \label{eq:t2mloss}
\end{equation}
where $\boldsymbol{\ell}^{(j)}$ are the predicted logits for patch $j$ and
$\mathbf{c}$ are the indices produced by the frozen \tok encoder
$E_{\text{VQ}}$. Posing language-guided motion as classification, rather than regression onto a
continuous, multimodal flow field, sidesteps mode-averaging and lets the model
represent ambiguous prompts as a distribution over codes.

\subsection{Wan motion adapter}
\label{sec:adapter}

\paragraph{Design principles.}
The adapter must add motion controllability to Wan~2.2 without disturbing the
pretrained prior, cheaply and reversibly. Three principles follow.
\emph{(i)~Zero-gated initialization:} the adapter contributes exactly zero at the
start of training, so the backbone generates as if unconditioned and the motion
pathway is admitted only as it earns gradient. \emph{(ii)~Non-invasive
injection:} adapters attach through forward hooks, so no line of the backbone is
edited and the mechanism transfers to other DiT implementations.
\emph{(iii)~Parallel cross-attention:} conditioning is added \emph{alongside},
not in place of, the block's computation, leaving the original flow intact.

\paragraph{Per-block adapter.}
Let $\mathbf{X}\in\mathbb{R}^{B\times L_{\text{vis}}\times 1536}$ be a block's
visual hidden state and $\mathbf{Z}_q\in\mathbb{R}^{B\times 64\times 128}$ the
motion embeddings. The adapter first lifts the motion tokens to the backbone
width with a linear map $\mathbf{M}=\mathbf{Z}_q\mathbf{W}_m\in\mathbb{R}^{B\times 64\times 1536}$,
layer-normalizes both streams, and runs multi-head cross-attention in which the
visual tokens query the motion tokens:
\begin{equation}
  \mathbf{A} = \operatorname{MHA}\!\big(
    Q{=}\tilde{\mathbf{X}},\;
    K{=}\tilde{\mathbf{M}},\;
    V{=}\tilde{\mathbf{M}}
  \big),
  \label{eq:crossattn}
\end{equation}
with $H{=}8$ heads of dimension $192$. Each visual token thus attends over all
$64$ motion tokens. No patch positional encoding is added to the adapter memory in this formulation. Consequently, jointly permuting its keys and values leaves the attention output unchanged: the adapter consumes a set of motion embeddings, not an explicitly aligned spatial field. Positional encodings in \ttm assign its output grid but do not survive the codebook lookup as distinct patch identities. Spatial allocation during generation must therefore depend on the backbone's image-conditioned visual features; we do not claim verified token-to-object grounding. The output is projected and fused back through a single
gated residual,
\begin{equation}
  \mathbf{X}' = \mathbf{X} + \tanh(\alpha)\cdot \mathbf{A}\mathbf{W}_o,
  \qquad \alpha \leftarrow 0,
  \label{eq:gate}
\end{equation}
At inference, a strength parameter $s$ multiplies this residual. The scalar gate $\alpha$ is learned per block and initialized to $0$, so
$\tanh(\alpha){=}0$ and $\mathbf{X}'{=}\mathbf{X}$ exactly at initialization.
Crucially, only the gate enforces the zero-output property: the projection
$\mathbf{W}_o$ is initialized normally, so that once the gate opens the adapter
has full-rank capacity rather than having to escape a degenerate zero
initialization.

\paragraph{Low-rank factorization.}
The adapter's parameter count is dominated by the dense projections of
\cref{eq:crossattn,eq:gate}, which map between the backbone width $1536$ and
itself. Since the motion memory is only $64$ tokens wide, these maps need not be
full rank, so we optionally factor every adapter projection
$\mathbf{W}\in\mathbb{R}^{1536\times 1536}$ as
$\mathbf{W}=\mathbf{W}_{\text{down}}\mathbf{W}_{\text{up}}$ with
$\mathbf{W}_{\text{down}}\in\mathbb{R}^{1536\times r}$,
$\mathbf{W}_{\text{up}}\in\mathbb{R}^{r\times 1536}$, and $r{=}256$. The
zero-gated structure of \cref{eq:gate} is unaffected, since the gate rather than
the projection enforces the identity at initialization, and the trainable adapter
drops from $1.14$B to $201$M parameters at essentially no cost in control
(\cref{tab:cost}).

\paragraph{Injection via forward hooks.}
A hook wraps each of the $30$ Wan~2.2 blocks: after a block returns its hidden
state, the hook applies the corresponding adapter with the clip's motion
embeddings and substitutes the result, so conditioning is woven through the full
depth of the DiT without any change to its source (\needle in
\cref{fig:pipeline}). The gate is learned per block, so the motion each admits,
$|\tanh(\alpha)|$, need not be uniform with depth; \cref{fig:pipeline} plots the
trained values, which vary rather than saturating.

\paragraph{Training objective.}
We train only the adapters, with the same rectified-flow / flow-matching
objective used by the backbone~\cite{rectifiedflow,flowmatching}. Given a
VAE-encoded clean latent $\mathbf{x}_0$ and noise
$\boldsymbol{\epsilon}\sim\mathcal{N}(\mathbf{0},\mathbf{I})$, we draw
$t\sim\mathcal{U}[0,1]$, apply Wan's time-shift
$t' = 7t/(1+6t)$, and form the interpolant
$\mathbf{x}_{t'} = (1-t')\,\mathbf{x}_0 + t'\,\boldsymbol{\epsilon}$. The model
predicts the velocity field, supervised against the rectified-flow target:
\begin{equation}
  \mathcal{L}_{\text{adapter}} =
  \big\lVert
    v_\theta(\mathbf{x}_{t'},\, t',\, \mathbf{H},\, \mathbf{Z}_q)
    - (\boldsymbol{\epsilon}-\mathbf{x}_0)
  \big\rVert_2^2,
  \label{eq:adapterloss}
\end{equation}
where $v_\theta$ is the frozen Wan~2.2 DiT augmented with the adapters, and
$\mathbf{Z}_q$ are ground-truth motion embeddings from the frozen tokenizer.
During adapter training we condition on tokenizer targets rather than \ttm
predictions, so that adapter learning is not coupled to language-model error;
the two are composed only at inference. The Wan~2.2 DiT, its VAE, the UMT5 encoder, and \tok are all frozen;
\cref{tab:params} summarizes the parameter budget.

\section{Experiments}
\label{sec:experiments}

We evaluate the three claims of the paper: that \sys (i)~achieves text-driven \emph{motion control} without explicit motion input;
(ii)~\emph{largely maintains the measured generative quality} of the frozen backbone; and
(iii)~does so \emph{with limited additional forward compute and no backbone updates}. Because claim~(i)
concerns a three-stage pipeline, we additionally isolate \emph{which} stage the
control comes from rather than reporting only the end-to-end number
(\cref{sec:ladder}).

\subsection{Datasets and trajectory extraction}
We train on $\sim\!462$K real-world videos pooled from
Something-Something~v2 (SSV2)~\cite{ssv2} and Kinetics-400 (K400)~\cite{kinetics}.
SSV2 contributes hand-object interactions with template captions
(\eg, ``pushing \emph{[something]} from left to right''), which give clean,
unambiguous motion supervision; K400 contributes diverse action categories with
broader scene motion. For every clip we extract a dense motion field with
CoTracker3~\cite{cotracker3} offline: a $32{\times}32$ grid of $1024$ points
tracked over $49$ frames, stored as velocities and an initial position and
normalized to $[-1,1]^2$. We hold out $10\%$ of the data \emph{by action class}
rather than by video, so evaluation measures generalization to unseen motion
descriptions rather than memorization.

\subsection{Implementation details}
The backbone is the $5$B-parameter Wan~2.2 DiT~\cite{wan22} ($30$ blocks, width
$1536$, $8$ heads) with its native VAE and the UMT5~\cite{umt5} text encoder.
\tok uses a $512$-entry codebook of dimension $128$, trained $100$ epochs with
AdamW; \ttm is a width-$512$, $4$-layer, $8$-head transformer ($\sim\!28$M
parameters) trained $100$ epochs against frozen tokenizer targets; adapters are
trained $30$ epochs with AdamW ($\text{lr}{=}10^{-4}$, cosine decay) under
\cref{eq:adapterloss}. Text embeddings and VAE latents are pre-computed once so
per-step cost is dominated by a single frozen forward pass. Everything runs on
one H100 (80GB); \cref{tab:params} gives the parameter budget.

\ViMolaQual
\subsection{Metrics and protocol}
\label{sec:protocol}
Motion control is easy to overstate with a metric that drifts between tables, so
we fix one protocol and hold it everywhere. \emph{Directional accuracy} (DA)
re-extracts flow from a generated clip, takes the foreground as the top quartile
of flow magnitude, reduces its dominant direction to one of four bins, and
compares it against the direction named by the caption template. Bin count,
foreground mask, label source, and chance are identical in every table, chance is
$0.250$, and we print it in each caption. DA uses a fixed $300$-clip subset for the main comparison, ladder, and strength sweep; complex-motion splits have separate sizes reported in the supplementary material. Quality metrics and the main-table trajectory error use the full $5{,}000$-clip held-out sets. We report $95\%$ Wilson intervals, score a \emph{single}
configuration fixed before evaluation, and measure quality at the same $s$ as
control.

Alongside DA we report Traj.\ L2, the error between re-extracted and conditioning
trajectories, and $\rho_{\text{spd}}$, the rank correlation between conditioning
and realized speed; for quality, FVD~\cite{fvd} on I3D~\cite{i3d} features and a
Motion Smoothness score; for alignment and identity, CLIP/ViCLIP~\cite{clip,viclip},
DINO subject and background consistency~\cite{dino}, and the frame-$0$ $L_2$
against the conditioning frame (\cref{sec:supp_protocol}). At $s{=}0.06$ the
gates settle to a mean $\tanh(\alpha)\!\approx\!0.028$, admitting motion as a
small, controlled residual.

\subsection{Baselines}
We evaluate two families end-to-end. The first takes, like \sys, no explicit
motion input: the frozen Wan~2.2~\cite{wan22} backbone with no adapter, run
image-to-video, isolating the adapter's contribution;
AnimateDiff+MotionLoRA~\cite{animatediff,lora}, run from a prompt alone; and
CogVideoX~\cite{cogvideox}, image-to-video, as a strong open reference. The
second is the natural alternative to our discrete design: predict a
\emph{continuous} trajectory field from the caption and hand it to a generator
that consumes trajectories. We build both halves, feeding a continuous regressor
into SG-I2V~\cite{sgi2v} and, separately, into our own adapter, so the discrete
bottleneck is compared against a continuous one under identical text-only
conditions. Controllers requiring \emph{ground-truth} motion at
inference~\cite{motionctrl,draganything,tora} solve a different problem
(\cref{sec:related}).

\subsection{Main results}
\cref{tab:main} reports the primary comparison. \sys reaches a directional
accuracy of $0.442$ against chance $0.250$, well ahead of the frozen backbone it
conditions ($0.287$) and of CogVideoX ($0.301$), and the lowest Traj.\ L2 error
($0.734$ against $0.812$, $0.789$, and $0.820$), despite receiving no explicit
tracking input. It also attains the highest CLIP score ($0.229$ \vs $0.221$, which we
read as parity rather than a win) and the lowest FVD on SSV2 ($168.9$ \vs
$179.4$), while on K400 the unmodified backbone retains a lower FVD ($320.3$ \vs
$328.5$) and higher smoothness ($0.472$ \vs $0.455$), a trade-off consistent with
conditioning on a signal it was not trained with.
AnimateDiff is the only method whose FVD is worse on SSV2 than K400 ($971.7$ \vs
$454.4$), which we attribute partly to its lacking a starting frame and partly to
its motion LoRA transferring poorly outside its tuning distribution; \sys shows
no such asymmetry.

\TableMain

The supplementary quality table reports overlapping intervals for SSV2 FVD ($168.9\,[161,177]$ versus $179.4\,[172,188]$) and DINO subject consistency ($0.731\,[.712,.749]$ versus $0.743\,[.725,.760]$). Background consistency is $0.812$ versus $0.821$; frame-$0$ $L_2$ is approximately $10^{-4}$ for both. These measurements support similar average quality, rather than proving identity preservation on every clip. Over three seeds, DA is $0.442{\pm}0.011$ versus $0.287{\pm}0.009$. A blind pairwise study agrees: seven raters judging $150$
matched-seed pairs chose \sys on direction $68.4\%$ of the time and on speed
$61.2\%$, split evenly on overall quality at $49.1\%$, and matched automatic DA
on $74.2\%$ of pairs (\cref{tab:human}).

\subsection{Where the control comes from}
\label{sec:ladder}
An end-to-end gain does not by itself show that the \emph{predicted codes}
produced it; a conditioning pathway can help simply by being open. We therefore
hold the trained adapter and the $300$ prompts fixed and vary only what the
adapter is conditioned on (\cref{tab:ladder}, top). Shuffled codes, drawn from
the same codebook but paired with the wrong prompt, land at $0.283$,
close to removing the adapter entirely ($0.287$). Predicted codes reach $0.442$, a gap of $+0.159$ over shuffled, supporting the importance of prompt-matched token content. Traj.\ L2 also improves, and $\rho_{\text{spd}}$ rises from $+0.04$ to $+0.34$. This shuffle exchanges codes across prompts; it does not test within-prompt spatial arrangement. Oracle codes reach $0.531$, $0.089$ above predicted codes. The remaining $0.469$ gap to perfect accuracy cannot be assigned solely to the tokenizer or backbone: tracking error, foreground selection, and caption ambiguity may also contribute. A real-video DA reference under the same protocol would be needed to separate these factors.

The continuous regressor in \cref{tab:ladder} reaches $0.352$ DA via SG-I2V~\cite{sgi2v} and $0.331$ via our adapter, versus $0.442$ for discrete codes. The corresponding FVD scores are $243.6$, $186.2$, and $168.9$. These comparisons support discrete motion prediction under both tested conditioning mechanisms. A matched text-only adapter baseline remains untested.

\TableLadder

\subsection{Cost and complex motion}
\label{sec:complex}
The full $1.14$B adapter adds $7.88\%$ to the FLOPs of a DiT forward pass, $0.31$\,s per clip ($+2.2\%$) at inference, and trains in $62.4$\,GB (\cref{tab:cost}). Factorizing its projections at rank $256$ reduces the trainable size to $201$M parameters, forward overhead to $1.45\%$, inference overhead to $0.09$\,s ($+0.6\%$), and training memory to $41.7$\,GB. DA is $0.428$ versus $0.442$ for the full adapter. Thus, the low-rank variant trades a small measured reduction in control for lower training and inference cost; the efficiency claim concerns additional compute, rather than a small absolute parameter count.

One code per patch for the whole clip is a deliberate simplification, and
\cref{tab:complex} quantifies it. On single-phase prompts \sys reaches $0.478$
against the backbone's $0.291$ ($+0.187$); on multi-phase prompts, split
automatically by a change in dominant flow direction, the gap narrows to
$+0.048$, and on compound prompts to $+0.054$. Control degrades but does
not vanish. A representation-level limitation is also visible: on the multi-phase split the tokenizer's own reconstruction error is
$0.147$ against $0.084$ elsewhere, and a temporal axis of $S$ sub-clip segments
reduces it to $0.104$ at $S{=}2$ and $0.089$ at $S{=}4$. These are tokenizer reconstruction results, not end-to-end generation results for a temporally segmented controller.

\subsection{Portability beyond Wan}
\label{sec:portability}
\cref{tab:backbone} tests whether the learned motion vocabulary can condition a second video generator. We reuse the tokenizer, its codebook, and \ttm unchanged, and retrain only the generator-side adapters for CogVideoX-5B-I2V. The text-to-motion component is therefore shared across the two backbones; the transfer does not require learning a second motion vocabulary or language predictor. It does require adapter training for the new generator, so this is not zero-shot transfer of the complete generation pipeline.

CogVideoX DA rises from $0.301$ without the adapter to $0.398$ with it, a gain of $+0.097$ at chance $0.250$. Wan~2.2 improves from $0.287$ to $0.442$ ($+0.155$). The positive gain on both supports reuse of the motion representation, while the smaller gain on CogVideoX shows that the effectiveness of conditioning still depends on the generator and its trained adapters. DINO subject consistency is $0.795$ versus $0.788$ on CogVideoX and $0.743$ versus $0.731$ on Wan. These point estimates accompany the control comparison; they do not establish exact equality of quality or backbone-independent performance.

\TableBackbone
\TableCost
\TableComplexMain
\TableAblation
\subsection{Interpreting low token accuracy}
\TableCodebookMain
Sweeping the codebook size $K$ (\cref{tab:codebookmain}) exposes the central
diagnostic for the pipeline. Hard Top-1 collapses by $4.3\times$ as the
vocabulary grows, from $22.1\%$ at $K{=}128$ to $5.1\%$ at $K{=}1024$, as the label space grows eightfold, yet every measurement of the \emph{decoded} motion
is flat or improves: end-point error moves only from $0.198$ to $0.187$ and DA
from $0.431$ to $0.445$. The divergence between token accuracy and decoded motion supports evaluating predictions in motion space, rather than treating every code mismatch as equally severe. At $K{=}512$, \ttm's Top-1 of $8.6\%$ therefore
coexists with a decoded end-point error of $0.188$ against $0.417$ for random and
$0.291$ for majority codes (\cref{tab:priors}), an $\varepsilon$-soft Top-1 of
$0.512$, and a mean cosine similarity of $0.71$ between decoded fields.
\cref{tab:t2m} explains the dataset difference: SSV2's templated captions are cleaner
supervision than K400's, yet SSV2's Top-1 is lower ($7.5\%$ \vs $15.9\%$),
because its codes have higher entropy ($5.77$ \vs $5.37$) and a far weaker
majority prior ($2.4\%$ \vs $12.7\%$) --- K400's scene motion is dominated by
global patterns a constant predictor captures. As lift over that prior the
ordering inverts, $3.1\times$ against $1.25\times$.

\subsection{Ablations}
The upper panel of \cref{tab:ablation} isolates two design choices of
\cref{sec:adapter}: \emph{the gate}, zero-gated $\tanh(\alpha)$ against an
ungated residual, testing the stability claim of \cref{eq:gate}; and
\emph{injection depth}, all $30$ blocks against a subset. The rest are ablated
elsewhere: the \emph{discrete bottleneck} in \cref{tab:ladder}, \emph{codebook
budget} in \cref{tab:codebook}, and \emph{adapter rank} in \cref{tab:cost}. The lower panel sweeps the inference-time knob $s$: DA climbs steeply from
$0.312$ at $s{=}0.01$ to $0.442$ at $s{=}0.06$ and then flattens ($0.451$ at
$s{=}0.10$), while quality turns the other way, FVD improving to $168.9$ before
degrading to $189.4$. We use $s{=}0.06$ as the pre-registered default.

\section{Limitations and Conclusion}
\label{sec:limitations}
\paragraph{Limitations.}
The adapter's attention is invariant to the order of its motion-memory tokens; a spatial tokenizer alone therefore does not establish spatial grounding in generation. \ttm also lacks image features, so counterfactual tests that relocate objects or introduce competing objects are needed to establish named-object control. The across-prompt shuffle tests content matching, and the continuous regressor tests one representation alternative; neither replaces within-prompt spatial tests or a matched text-only adapter baseline.

Oracle DA is $0.531$, and no real-video DA reference is available to disentangle tracking and label ambiguity from representation and generation error. One code per patch summarizes an entire clip: gains fall from $+0.187$ on single-phase prompts to about $+0.05$ on multi-phase and compound prompts. Temporal segmentation improves tokenizer reconstruction but remains untested end to end. Evidence for transfer covers two backbones, partly templated training captions, and fixed $480$px, $49$-frame generation.

\paragraph{Conclusion.}
\sys connects language-predicted motion tokens to frozen video generators through zero-gated adapters. Controlled and human evaluations support better directional control with similar measured quality. CogVideoX results support vocabulary reuse; spatial grounding and multi-phase control remain open challenges.

\clearpage
{
    \small
    \bibliographystyle{ieeenat_fullname}
    \bibliography{main}
}
\clearpage
\appendix
\onecolumn
\setcounter{figure}{0}
\renewcommand{\thefigure}{A\arabic{figure}}
\setcounter{table}{0}
\renewcommand{\thetable}{A\arabic{table}}
\renewcommand{\theHsection}{supp.\arabic{section}}
\renewcommand{\theHfigure}{supp.\arabic{figure}}
\renewcommand{\theHtable}{supp.\arabic{table}}
\begin{center}
{\Large\bfseries Supplementary Material}
\end{center}
%══════════════════════════════════════════════════════════════════════════════
\section{Pipeline Visualization}
 
\begin{figure}[!htbp]
  \centering
  \begin{subfigure}[t]{0.26\linewidth}
    \includegraphics[width=\linewidth]{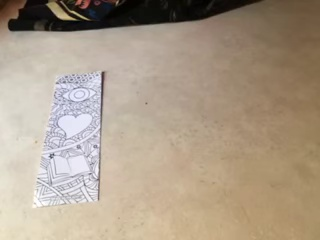}
    \caption{Input frame}
  \end{subfigure}\hfill
  \begin{subfigure}[t]{0.26\linewidth}
    \includegraphics[width=\linewidth]{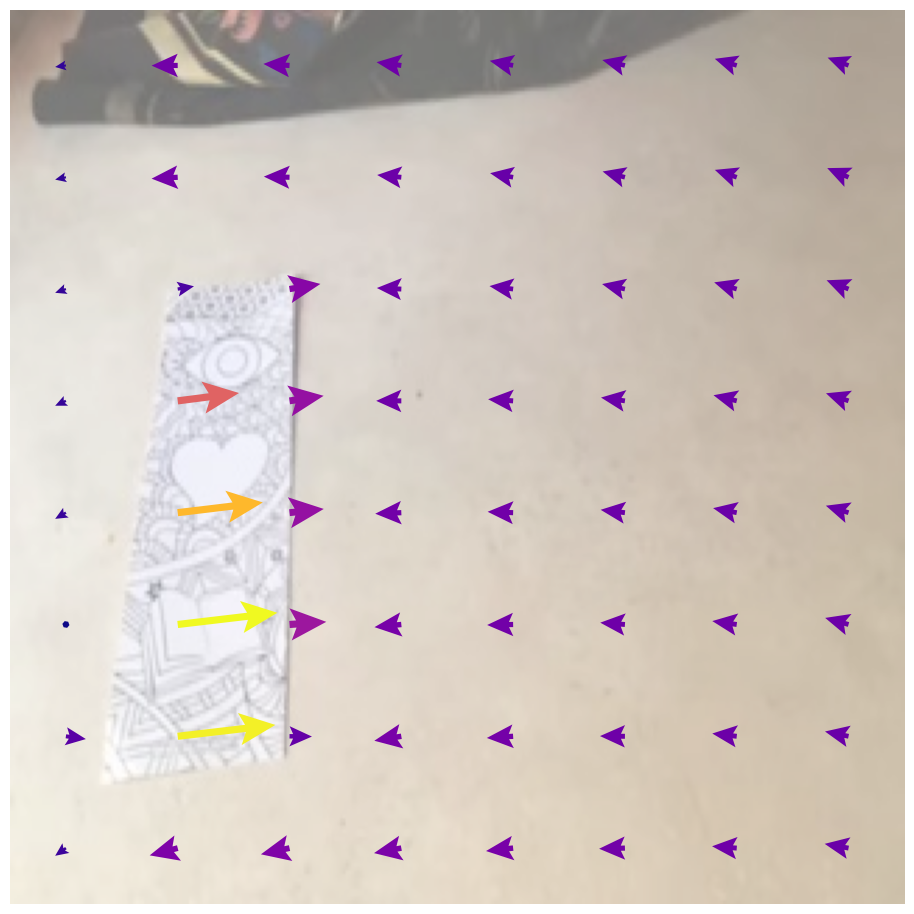}
    \caption{Dense motion field overlay}
  \end{subfigure}\hfill
  \begin{subfigure}[t]{0.26\linewidth}
    \includegraphics[width=\linewidth]{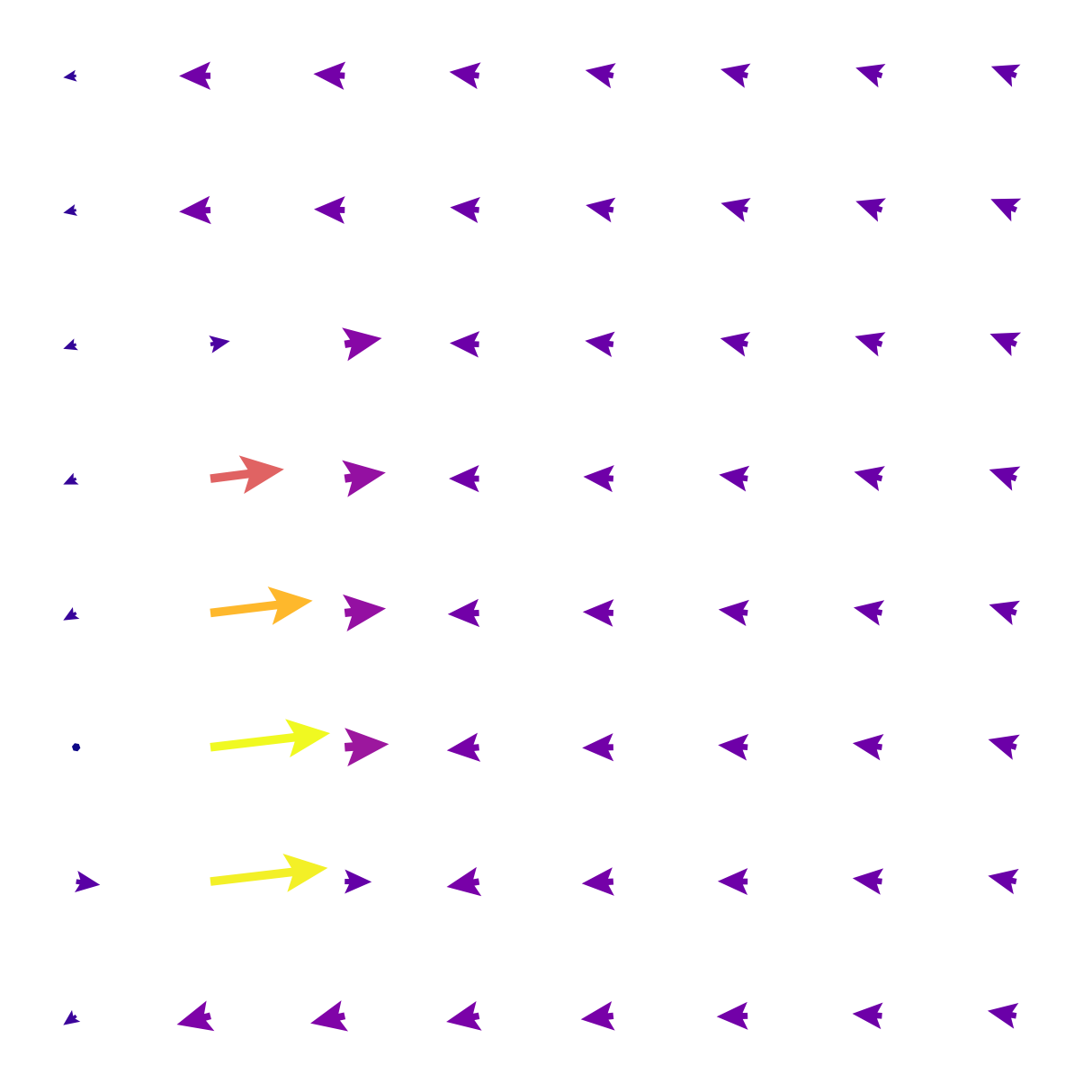}
    \caption{Velocity field}
  \end{subfigure}
 
  \vspace{3pt}
  \begin{subfigure}[t]{0.26\linewidth}
    \includegraphics[width=\linewidth]{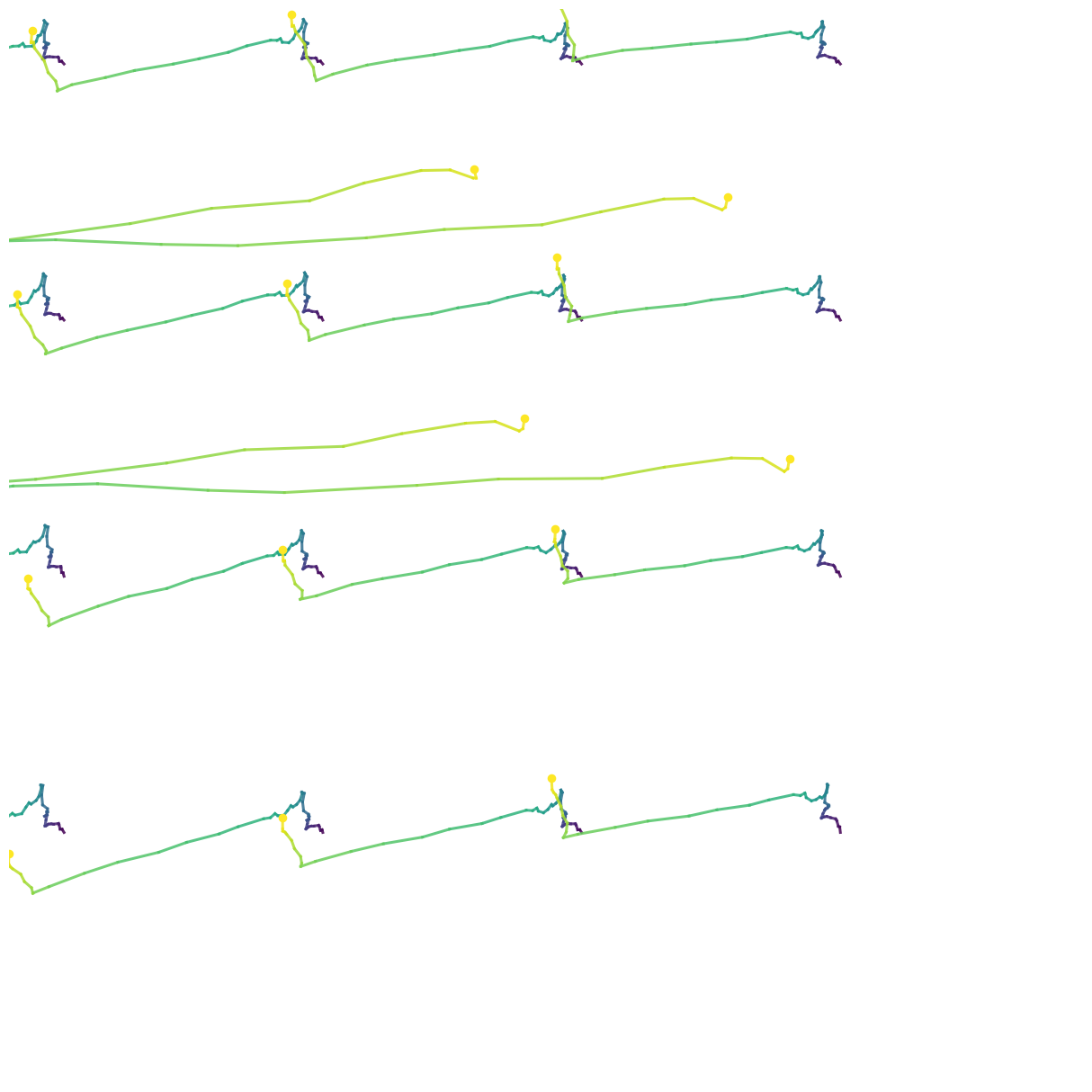}
    \caption{Trajectory traces ($49$ frames)}
  \end{subfigure}\hfill
  \begin{subfigure}[t]{0.26\linewidth}
    \includegraphics[width=\linewidth]{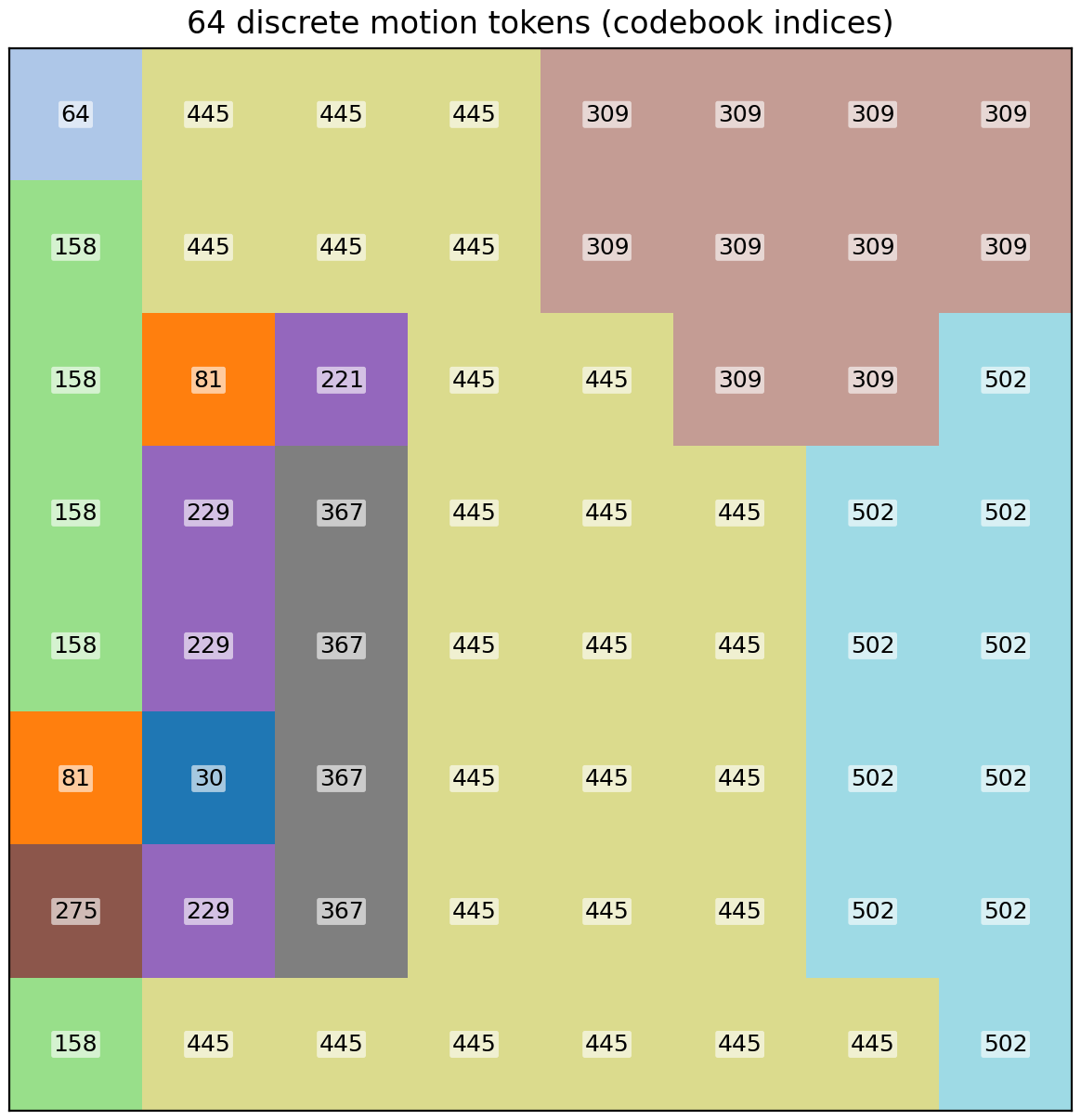}
    \caption{VQ token grid ($8{\times}8 = 64$ tokens)}
  \end{subfigure}\hfill
  \begin{subfigure}[t]{0.26\linewidth}
    \includegraphics[width=\linewidth]{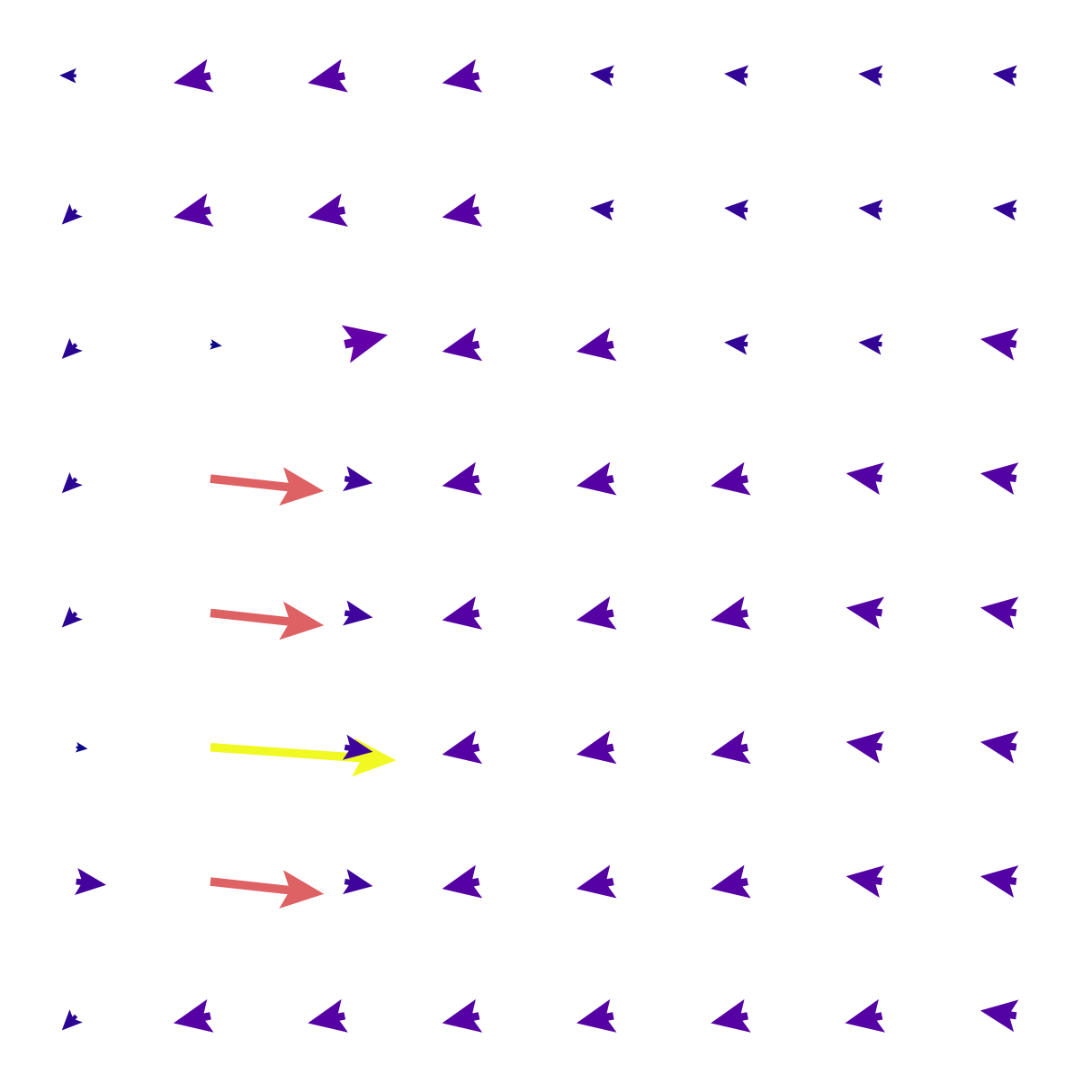}
    \caption{Reconstructed velocity field}
  \end{subfigure}
 
  \vspace{3pt}
  \begin{subfigure}[t]{0.45\linewidth}
    \includegraphics[width=\linewidth]{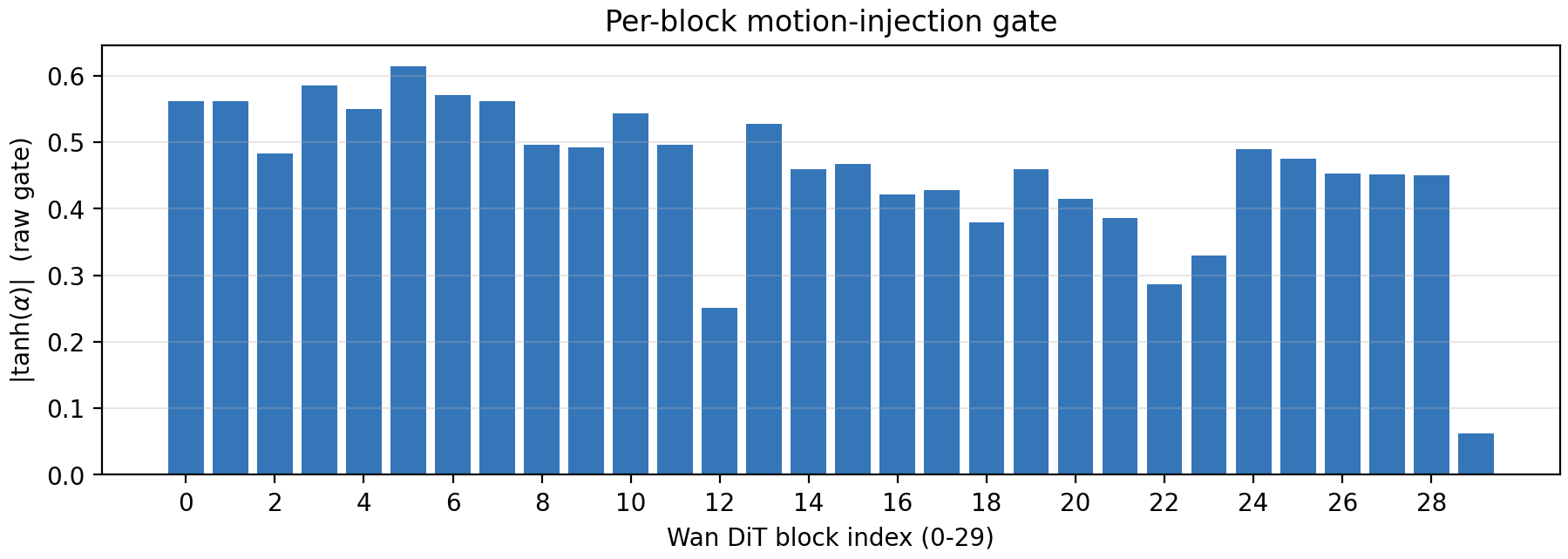}
    \caption{Per-block gate activations $\tanh(\alpha_i)$ after training}
  \end{subfigure}
 
  \caption{\textbf{Pipeline visualization.}
  (a) A raw SSV2 video frame.
  (b--c) The CoTracker3 motion field overlaid on and separated from the frame.
  (d) Trajectory traces integrated from the $32{\times}32$ tracked grid over $49$ frames.
  (e) The $8{\times}8 = 64$ discrete VQ token assignments produced by \tok
      (color = codebook index mod 32).
  (f) The velocity field reconstructed by the VQ decoder --- faithful to (c).
  (g) Per-block gate activation $\tanh(\alpha_i)$ across the $30$ DiT blocks after
      adapter training. All gates start at $0$; at default inference strength
      $s{=}0.06$ the mean activation is ${\approx}0.028$, confirming that the
      motion signal is injected as a small, controlled residual.}
  \label{fig:pipeline_supp}
\end{figure}
 
\clearpage
% ══════════════════════════════════════════════════════════════════════════════
\clearpage
\section{Qualitative Method Comparison}
 
Eight-frame temporal strips for 4 held-out SSV2 clips.
Each block shows, top to bottom:
\textbf{(1) \sys (ours)},
\textbf{(2) Wan~2.2} (no adapter),
\textbf{(3) CogVideoX-1.5},
\textbf{(4) AnimateDiff\,+\,MotionLoRA}.
\sys, Wan, and CogVideoX are image-to-video (TI2V) models conditioned on the
SSV2 first frame and text prompt; AnimateDiff is text-to-video only (no frame).

\paragraph{Observations.}
\textbf{CogVideoX-1.5}, despite being a strong TI2V generator, consistently
does not consistently follow the \emph{direction} or \emph{type} of motion described in
the caption.  It tends to synthesize plausible-looking generic motion
(\eg, a slow zoom or drift) that is semantically disconnected from the
action label --- the object may move, but not in the way the prompt specifies.

\textbf{AnimateDiff\,+\,MotionLoRA} suffers from a compounded failure: because
it receives no conditioning frame, it first loses visual and semantic fidelity
to the scene (\eg, wrong objects, backgrounds, or identities), and then also
fails to reproduce the described motion pattern.  The MotionLoRA prior biases
clips toward generic camera or body motions from its training distribution,
further decoupling the output from the action-specific trajectory implied by
the text.  As a result, AnimateDiff scores poorly on both CLIP alignment
\emph{and} trajectory-based metrics in our quantitative evaluation.

\sys and the Wan~2.2 baseline share the same first frame, so scene fidelity
is preserved across those two rows; the adapter's contribution is visible
primarily in the \emph{consistency} and \emph{directionality} of object motion
relative to the caption.

\begin{figure}[!htbp]\centering
  \includegraphics[width=\linewidth]{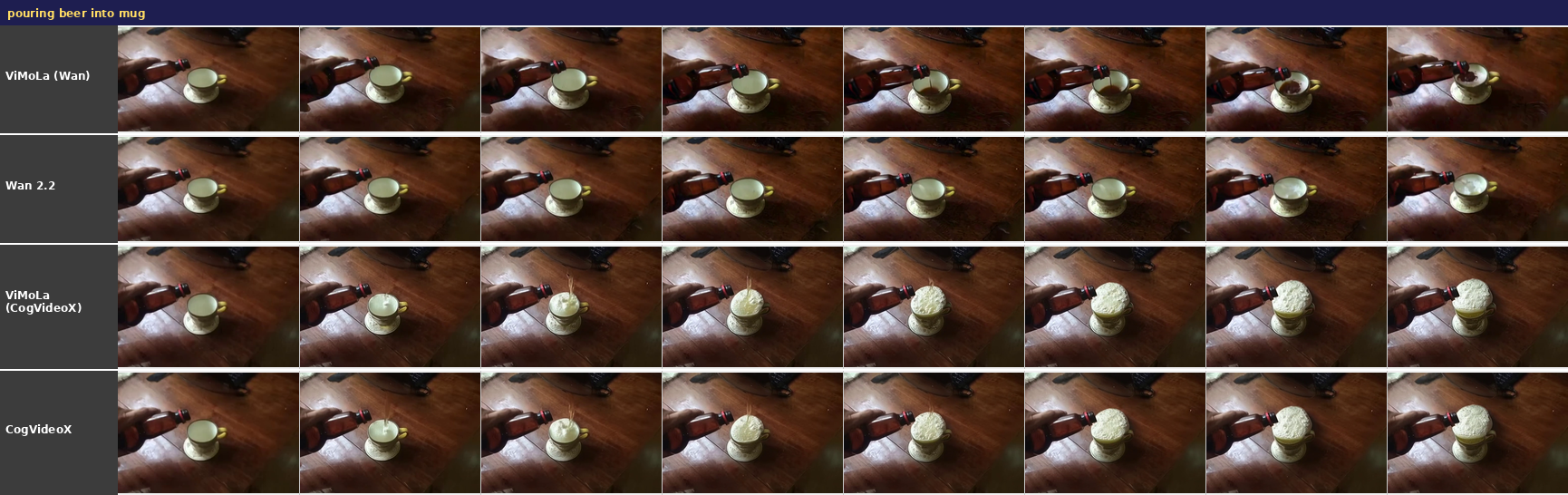}
  \caption{``pouring beer into mug''}
\end{figure}

\begin{figure}[!htbp]\centering
  \includegraphics[width=\linewidth]{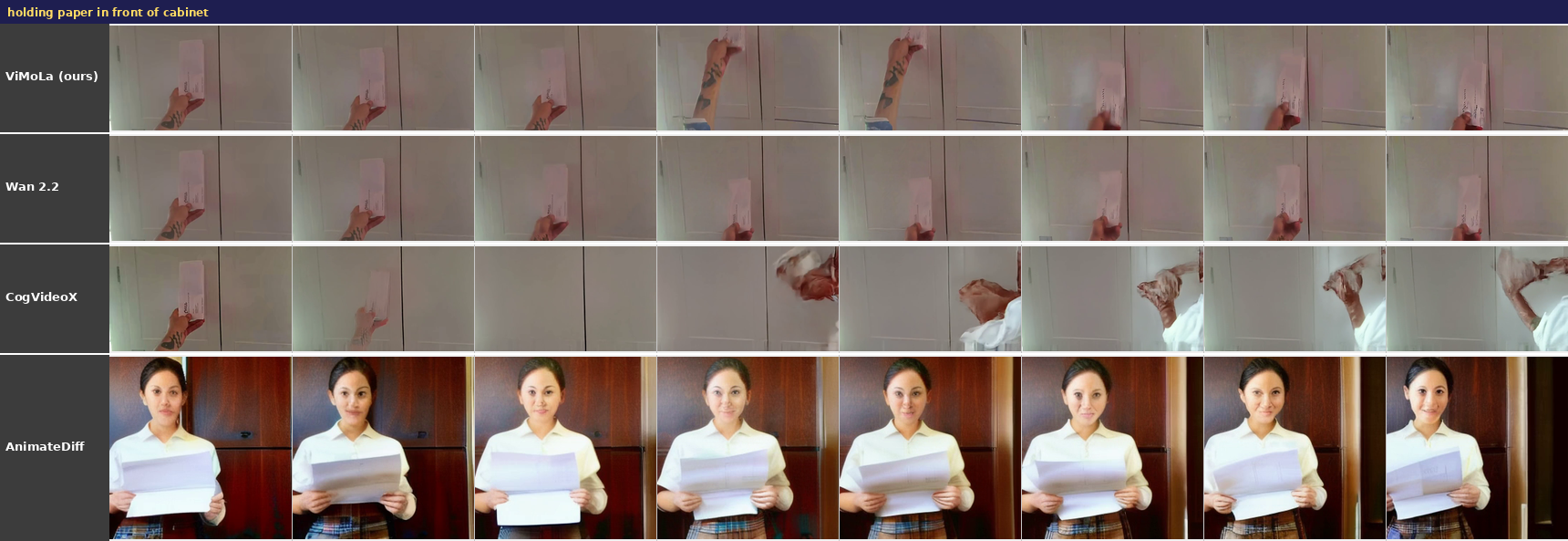}
  \caption{``holding paper in front of cabinet''}
\end{figure}
\begin{figure}[!htbp]\centering
  \includegraphics[width=\linewidth]{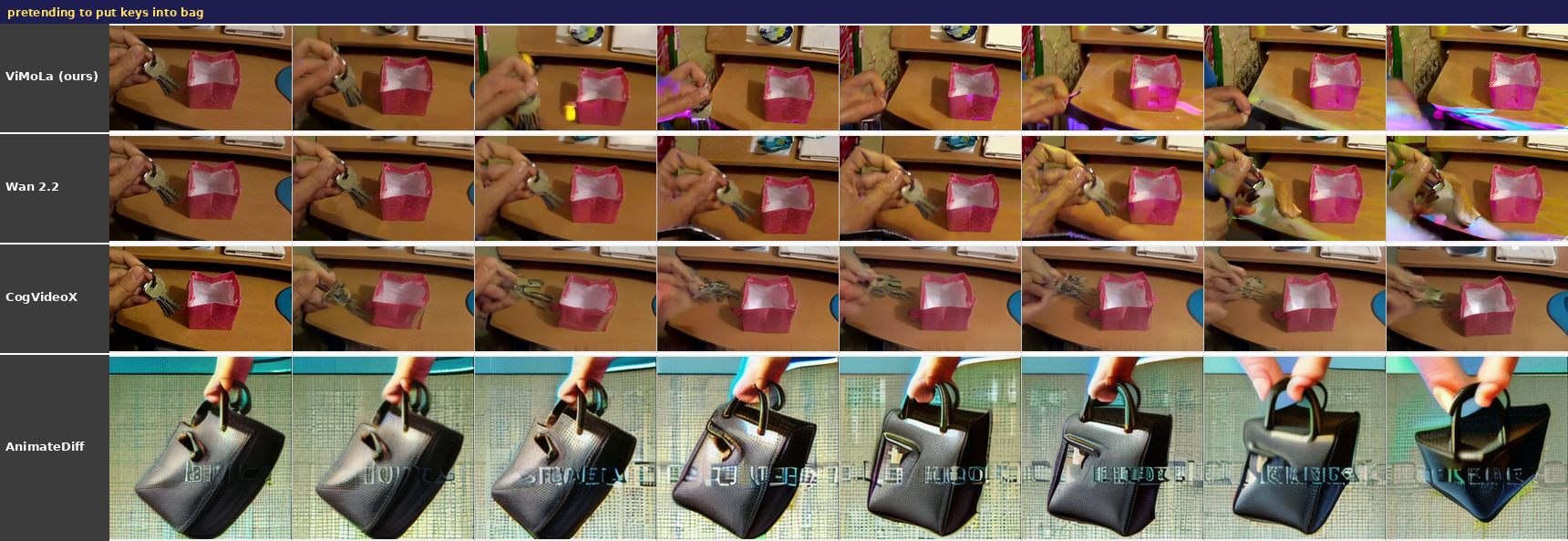}
  \caption{``pretending to put keys into bag''}
\end{figure}

\begin{figure}[!htbp]\centering
  \includegraphics[width=\linewidth]{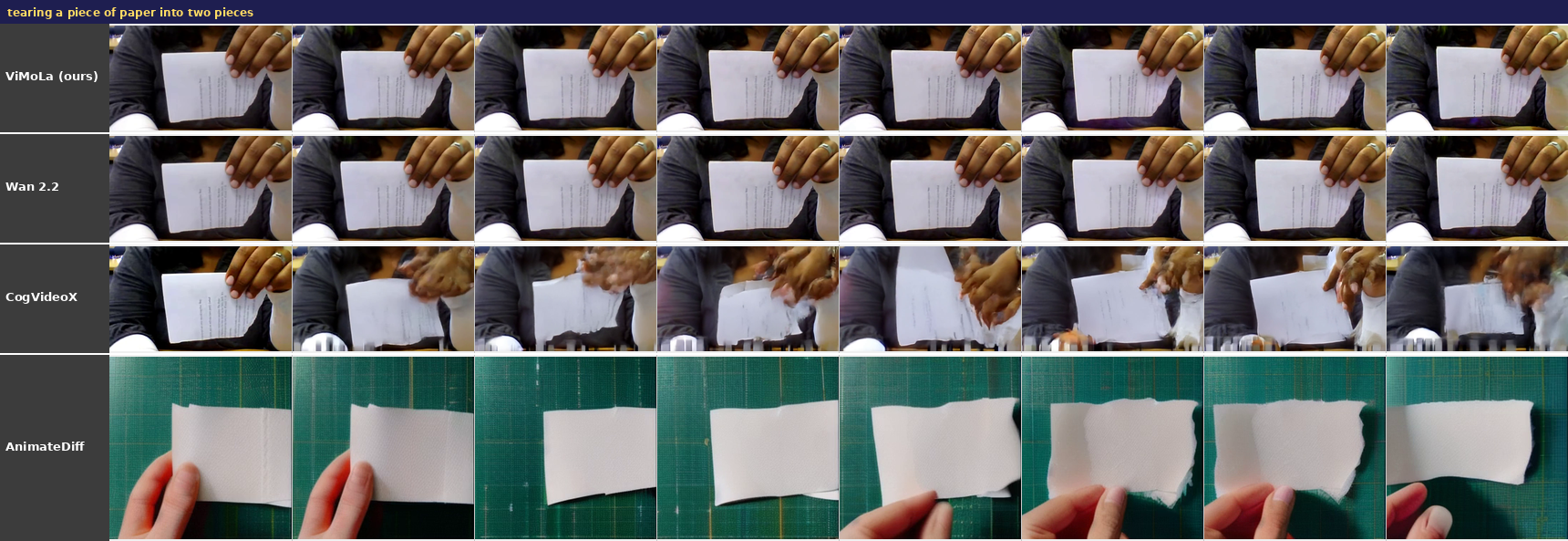}
  \caption{``tearing a piece of paper into two pieces''}
\end{figure}

\clearpage
% ══════════════════════════════════════════════════════════════════════════════
\clearpage
\section{Ground-Truth Motion Fields}
 
Each panel shows the SSV2 first frame (left) and the ground-truth CoTracker3
trajectory traces (right): $256$ of the $1024$ tracked points ($32{\times}32$
grid) drawn over $49$ frames, colored by total displacement magnitude
(plasma colormap, brighter = more motion).
These fields train \tok; their quantized indices are classification targets
for \ttm; they are \emph{not} available at inference.
 
% ── row 1 (original 8) ───────────────────────────────────────────────────────
\begin{figure}[!htbp]
  \begin{subfigure}[t]{0.42\textwidth}
    \includegraphics[width=\textwidth]{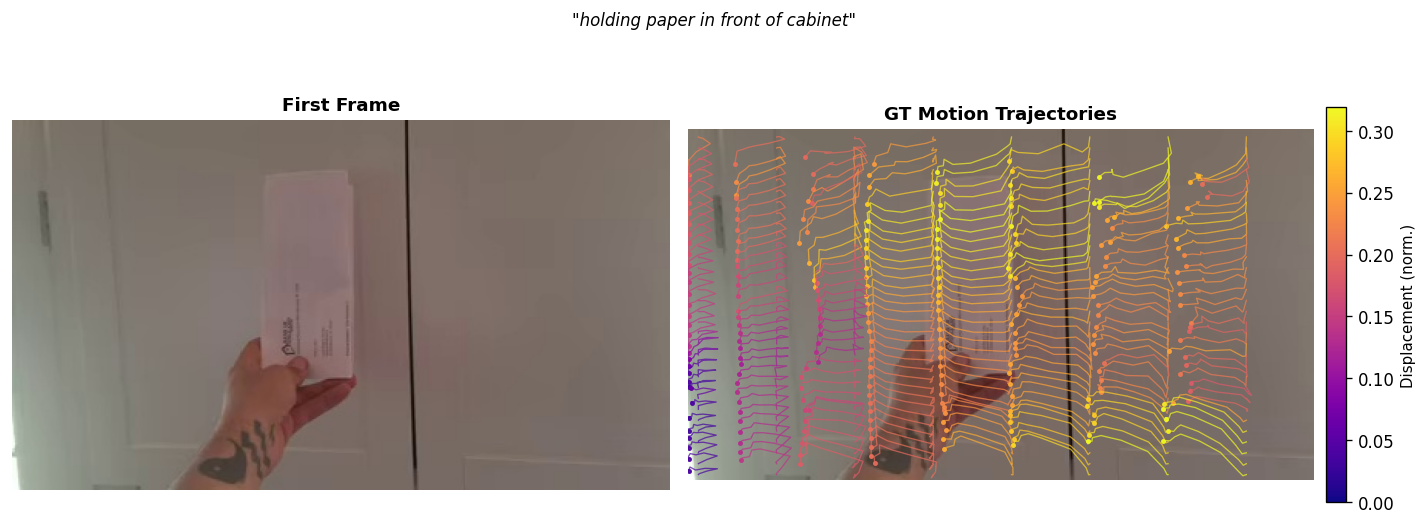}
    \caption{``holding paper in front of cabinet''}
  \end{subfigure}\hfill
  \begin{subfigure}[t]{0.42\textwidth}
    \includegraphics[width=\textwidth]{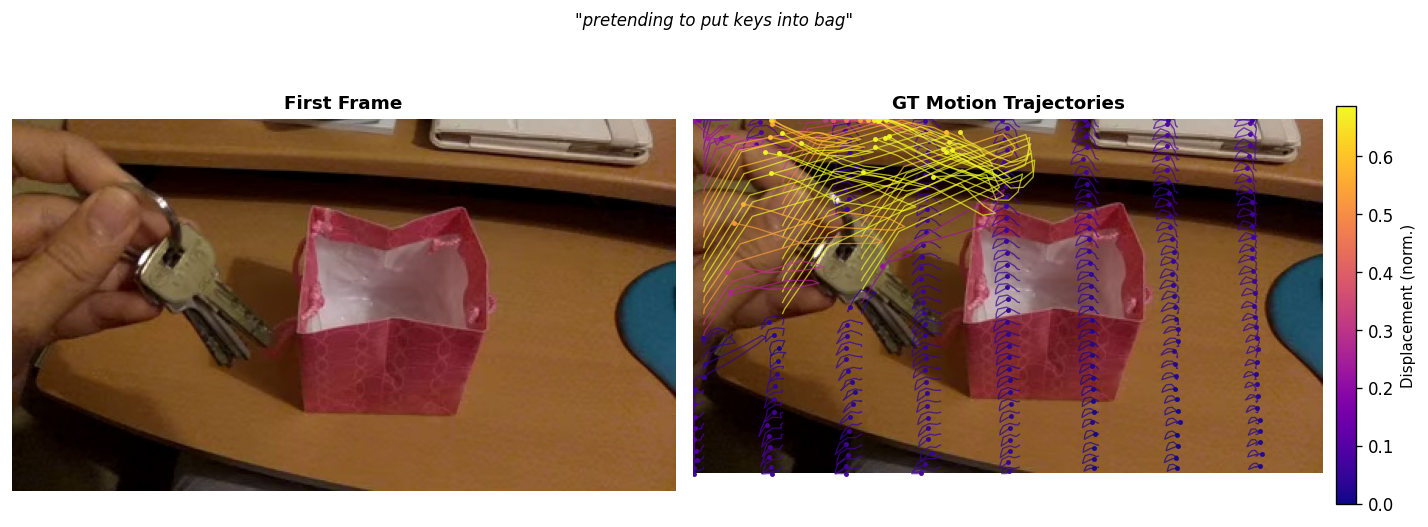}
    \caption{``pretending to put keys into bag''}
  \end{subfigure}
\end{figure}
 
\begin{figure}[!htbp]
  \begin{subfigure}[t]{0.42\linewidth}
    \includegraphics[width=\linewidth]{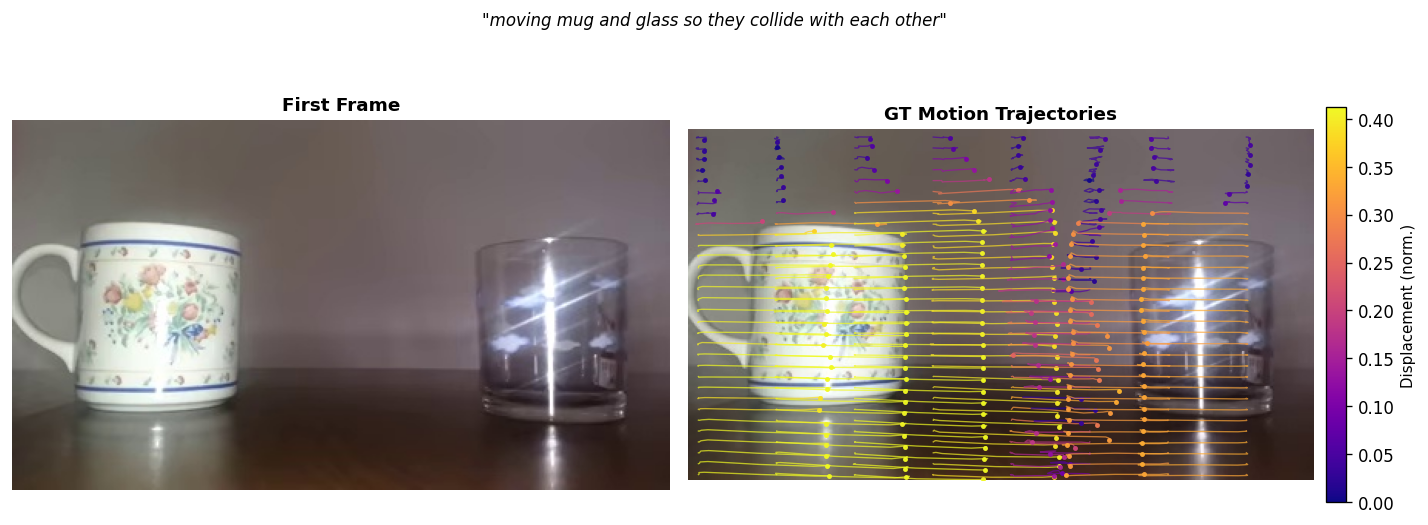}
    \caption{``moving mug and glass so they collide''}
  \end{subfigure}\hfill
  \begin{subfigure}[t]{0.42\linewidth}
    \includegraphics[width=\linewidth]{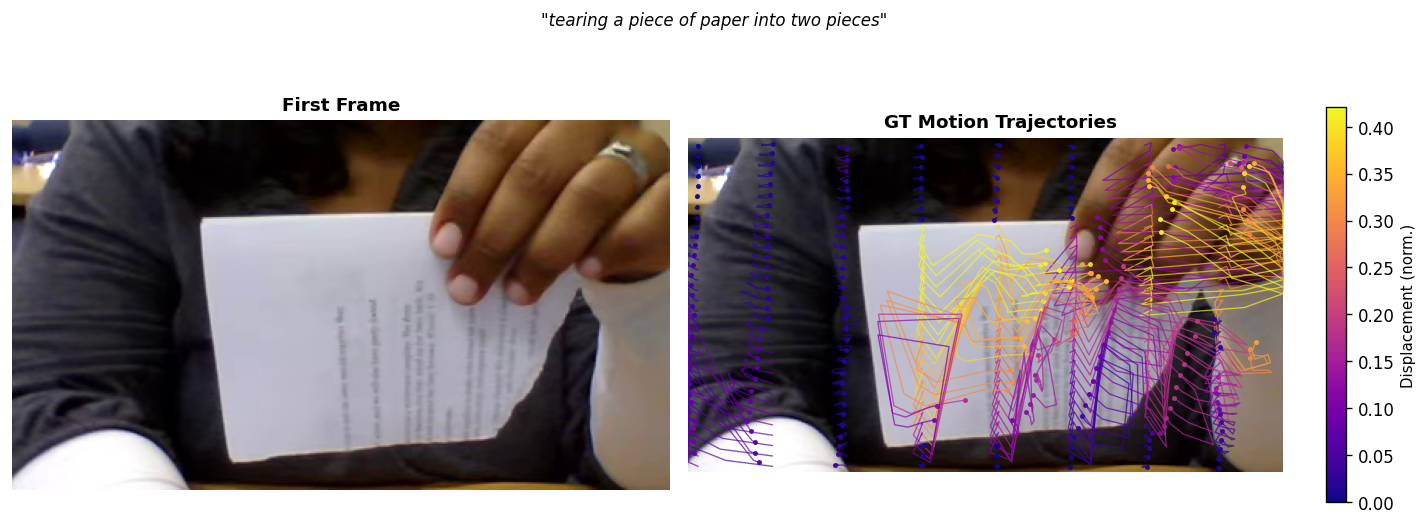}
    \caption{``tearing a piece of paper into two pieces''}
  \end{subfigure}
\end{figure}

% ── row 2 (additional 10) ────────────────────────────────────────────────────
\begin{figure}[!htbp]
  \begin{subfigure}[t]{0.42\linewidth}
    \includegraphics[width=\linewidth]{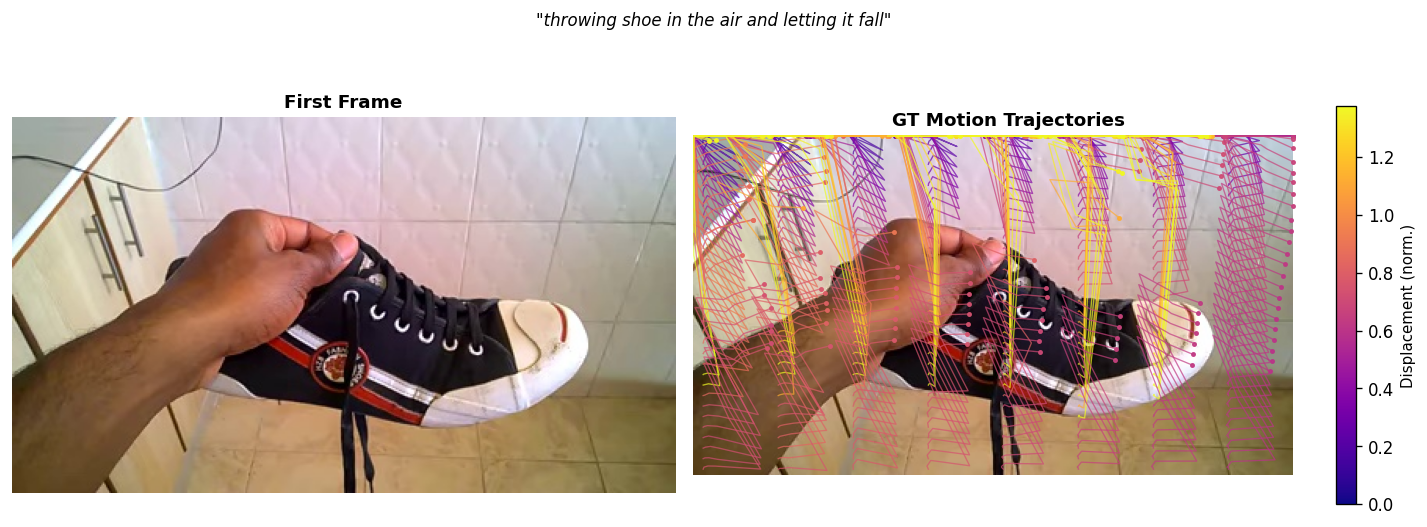}
    \caption{``throwing shoe in the air and letting it fall''}
  \end{subfigure}\hfill
  \begin{subfigure}[t]{0.42\linewidth}
    \includegraphics[width=\linewidth]{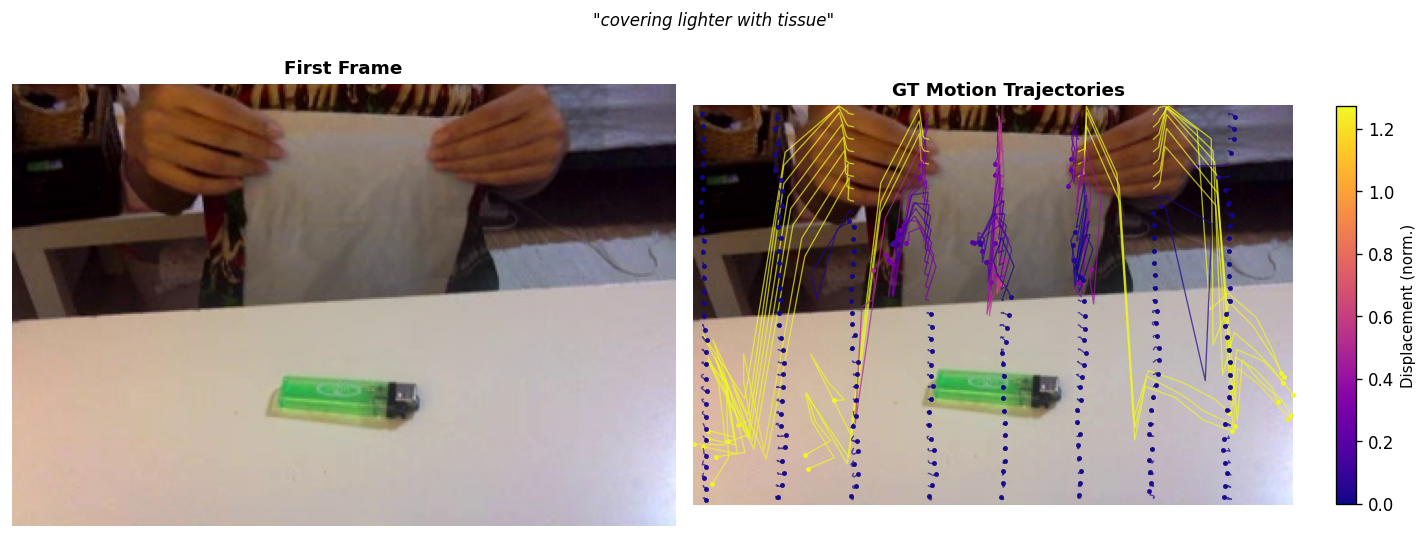}
    \caption{``covering lighter with tissue''}
  \end{subfigure}
\end{figure}
 
\begin{figure}[!htbp]
  \begin{subfigure}[t]{0.42\linewidth}
    \includegraphics[width=\linewidth]{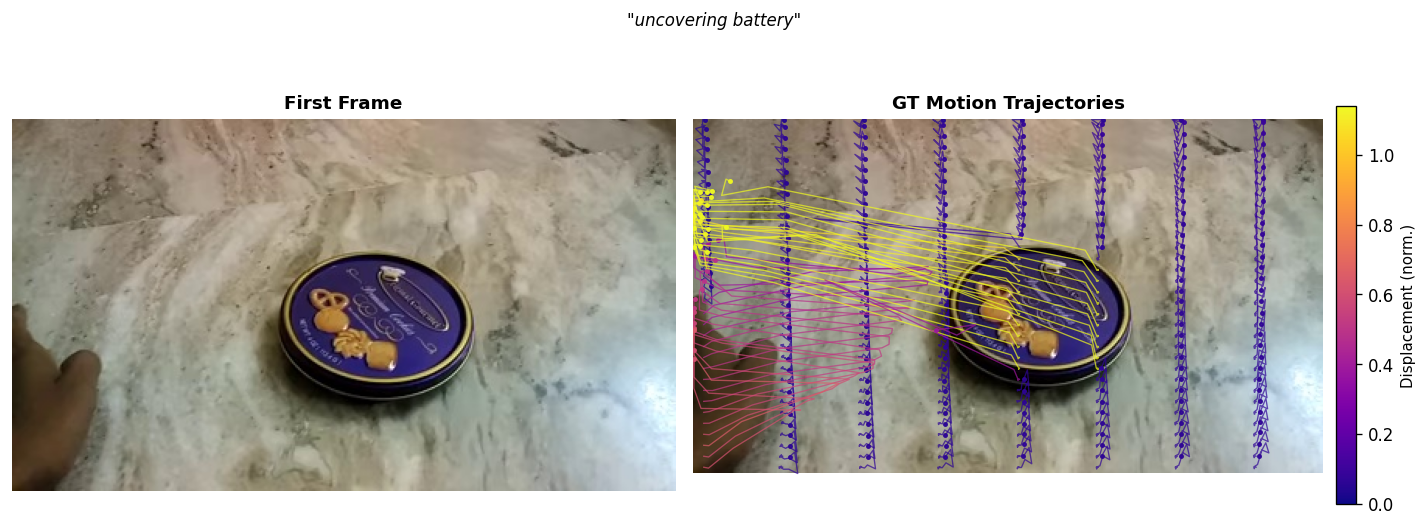}
    \caption{``uncovering battery''}
  \end{subfigure}\hfill
  \begin{subfigure}[t]{0.42\linewidth}
    \includegraphics[width=\linewidth]{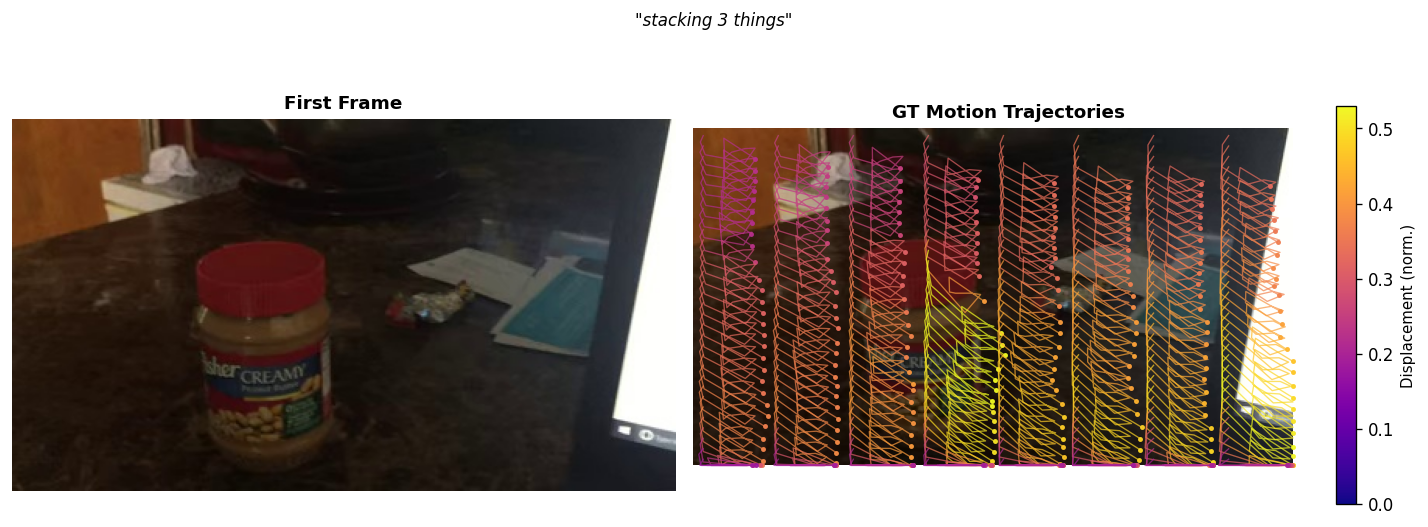}
    \caption{``stacking 3 things''}
  \end{subfigure}
\end{figure}

\begin{figure}[!htbp]
  \begin{subfigure}[t]{0.42\linewidth}
    \includegraphics[width=\linewidth]{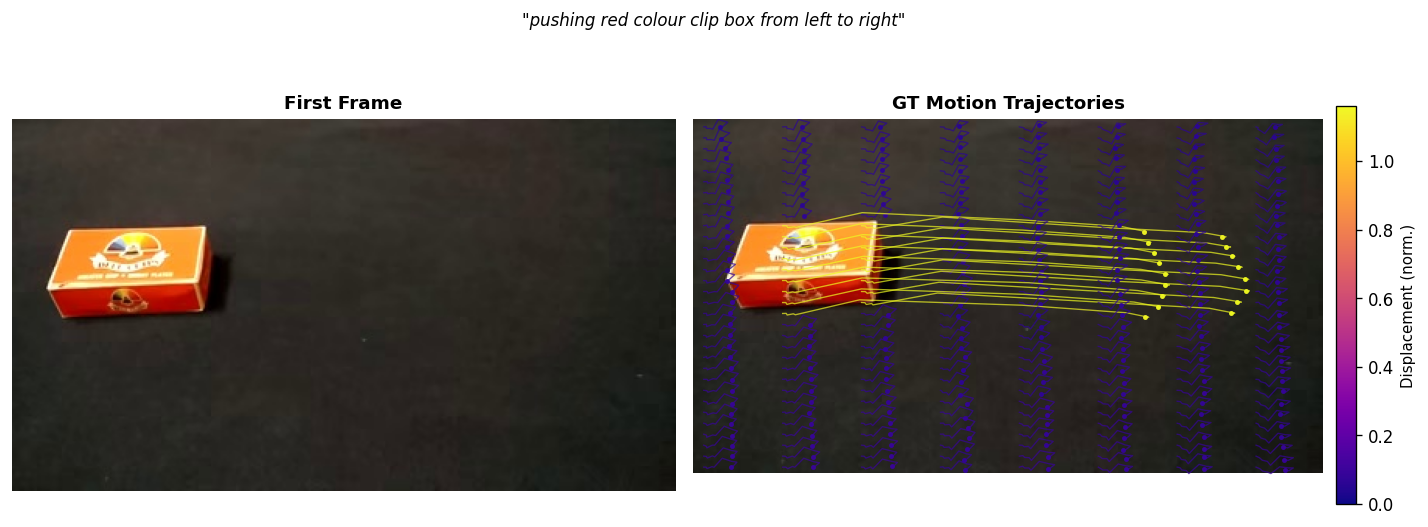}
    \caption{``pushing red colour clip box from left to right''}
  \end{subfigure}\hfill
  \begin{subfigure}[t]{0.42\linewidth}
    \includegraphics[width=\linewidth]{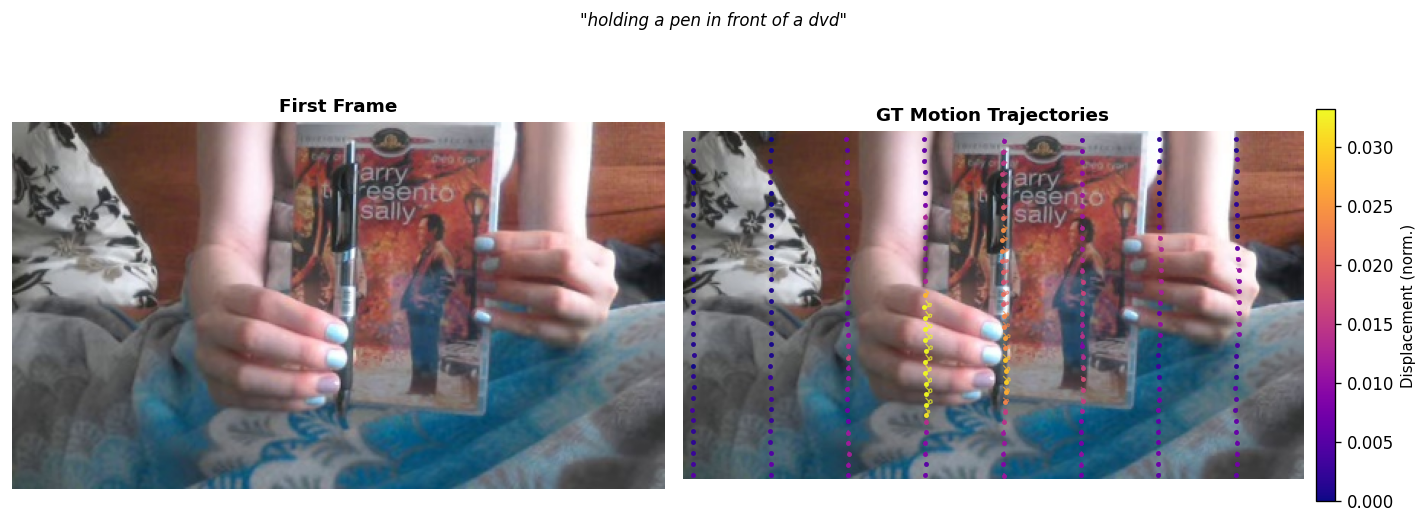}
    \caption{``holding a pen in front of a dvd''}
  \end{subfigure}
\end{figure}
 
% ── references ────────────────────────────────────────────────────────────────

\clearpage
\section{Motion From Text Alone vs.\ a Trajectory-Conditioned Controller}
\label{sec:supp_sisyphus}
\begin{figure}[!htbp]\centering
  \includegraphics[width=0.62\linewidth]{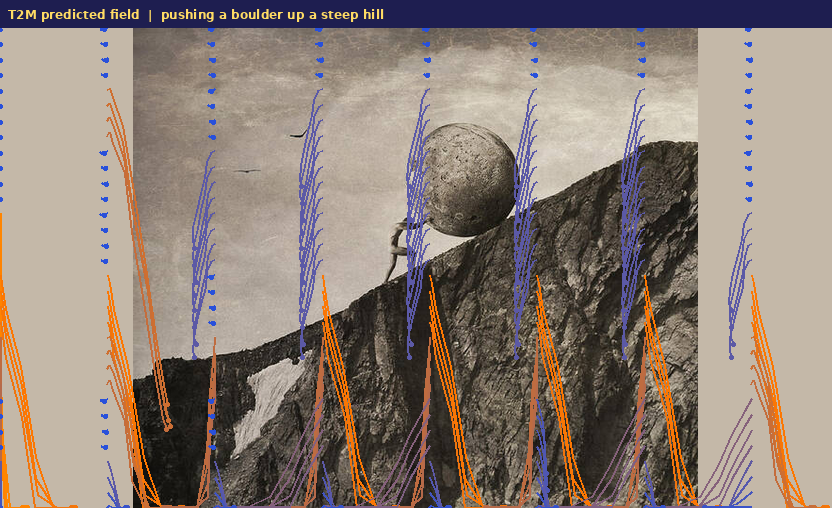}
  \caption{\sys's \ttm field decoded from ``pushing a boulder up a steep hill.''
  Predicted from language alone; no tracks are supplied at any point.}
  \label{fig:sisyphus_field}
\end{figure}

\begin{figure}[!htbp]\centering
  \includegraphics[width=\linewidth]{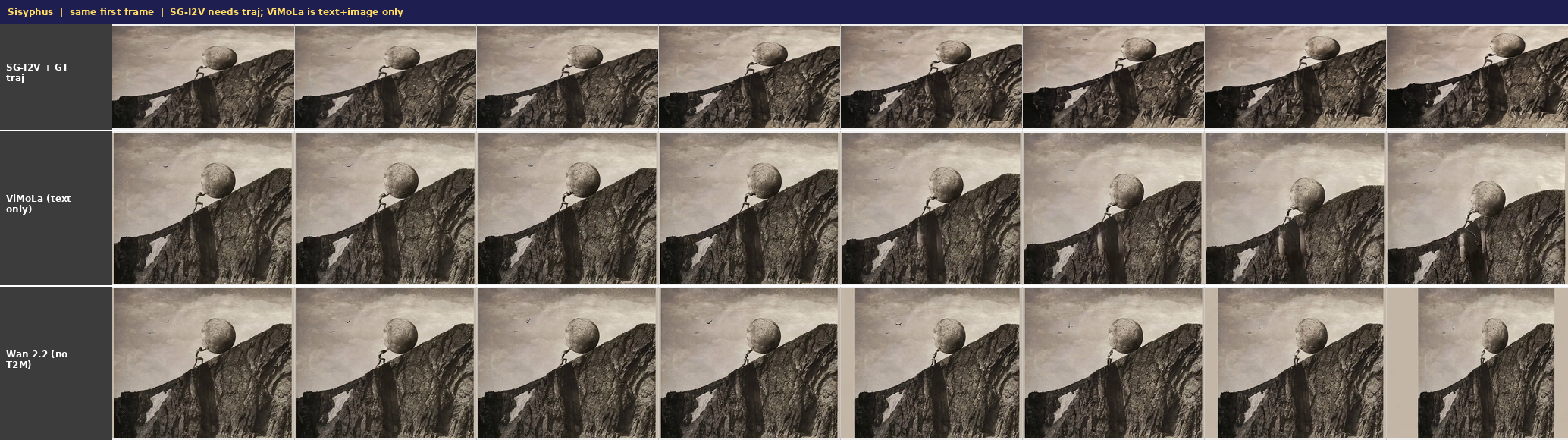}
  \caption{\textbf{Top:} SG-I2V, driven by ground-truth trajectories supplied at
  inference. \textbf{Middle:} \sys, conditioned only on the first frame and the
  caption, through the predicted field of \cref{fig:sisyphus_field}.
  \textbf{Bottom:} Wan~2.2, same prompt and frame with the adapter off. This is
  an out-of-distribution artistic image, shown at $s{=}0.1$ rather than the
  default $0.06$ (main-paper Table~6, lower panel).}
  \label{fig:sisyphus}
\end{figure}

All three rows of \cref{fig:sisyphus} start from the same first frame.
\textbf{SG-I2V}~\cite{sgi2v} is handed ground-truth boxes and centre paths at
inference; \sys receives no motion input at all, decoding a discrete field from
the caption (\cref{fig:sisyphus_field}) and injecting it into the frozen
backbone; Wan~2.2 is the same prompt and frame with the adapter off. Because
SG-I2V is given motion at test time, this is complementary to main-paper Table~2
rather than a substitute for it.

\noindent The middle row is the one with no motion signal, and also the one whose scene
survives the motion imposed on it: the ridge keeps its slope and silhouette, the
boulder holds its scale and its contact with the figure, and the displacement
accumulates smoothly. The other two each give something up. SG-I2V moves in the
right direction, but with supplied trajectories, and the
ridge deforms as the sequence advances. Wan~2.2 supplies almost no directed
motion; its geometry drifts, and by the final frame the slope has flattened and
the framing has shifted. In this example, \sys produces directed motion from language while maintaining more of the visible scene structure. This qualitative comparison does not establish object-specific grounding.

\clearpage
% ══════════════════════════════════════════════════════════════════════════════
\clearpage
\section{Evaluation Protocol Details}
\label{sec:supp_protocol}

\paragraph{Directional accuracy.}
Every DA number in the paper is computed identically. We re-extract dense
optical flow from the generated clip, define the foreground as the points whose
flow magnitude falls in the top quartile, average the foreground flow over the
clip to obtain a single dominant direction, and quantize that direction into one
of four bins (left, right, up, down). The prediction is scored against the
direction named by the caption \emph{template}, never against anything read off
the generated clip, so the label source is independent of the model under test.
With four equiprobable bins, chance is $0.250$; we print this value in every
caption that reports DA so that no accuracy figure has to be interpreted in
isolation. Intervals on DA are $95\%$ Wilson intervals for a binomial
proportion, and comparisons between two DA values use a two-proportion $z$-test.

\paragraph{Sample sizes and configuration count.}
The main DA comparison, ladder, and strength sweep use a fixed subset of $300$ held-out clips. Ladder trajectory and speed diagnostics use this same subset; main-paper Table~1 trajectory error uses $5{,}000$ clips per dataset. Complex-motion diagnostics use the separate split sizes reported in \cref{tab:complex}. Distributional quality metrics (FVD) use the full
$5{,}000$-clip held-out sets, since FVD is unreliable at small $n$. Codebook and
text-to-motion diagnostics, which require no video generation, use $4{,}000$
validation clips. We evaluated a single adapter configuration, fixed before
scoring, and report quality metrics at the same inference strength $s{=}0.06$ as
the control metrics; main-paper Table~6, lower panel is the only panel in which $s$ varies,
and it varies control and quality together.

\paragraph{$\varepsilon$-soft Top-1 and $\rho_{\text{spd}}$.}
The $\varepsilon$-soft Top-1 of \cref{tab:codebook} counts a predicted code as
correct when the velocity field decoded from it lies within $\varepsilon$ of the
field decoded from the ground-truth code, with $\varepsilon$ set to the
tokenizer's own reconstruction error at that codebook size. The threshold is
therefore calibrated rather than chosen: a prediction counts as correct exactly
when substituting it for the true code changes the decoded motion by no more
than the tokenizer already does. $\rho_{\text{spd}}$ is the rank correlation,
across the evaluation subset, between the mean speed of the conditioning motion
field and the mean speed re-extracted from the generated clip.

\TableParams
\ifdefined\StandaloneSupplement\TableCost\fi
\TableMotion
\TableQuality
\TableComplex
\ifdefined\StandaloneSupplement\TableBackbone\fi
\TableCodebook
\TableTokenizerDim
\TablePriors

% \clearpage
% ══════════════════════════════════════════════════════════════════════════════
\clearpage
\section{Human Study Details}
\label{sec:supp_human}

Seven raters each judged $150$ clip pairs, giving $1{,}050$ judgments in total.
Each pair contains one clip from \sys and one from the frozen Wan~2.2 backbone,
generated from the same prompt, the same conditioning frame, and the same random
seed, so that the two clips differ only in whether the adapter was attached.
Presentation order within each pair was randomized, and raters were not told
which system produced which clip or that one of them was the unmodified
baseline. For each pair a rater answered three independent questions: which clip
moves in the direction the caption describes, which clip moves at the speed the
caption implies, and which clip looks better overall. Ties were not permitted on
the first two questions and were split evenly on the third.

\TableHuman

Results are in \cref{tab:human}. The separation between the direction and
speed questions ($68.4\%$ and $61.2\%$) and the overall-quality question
($49.1\%$, an interval containing $50\%$) is the point of the study: raters see
the motion change and do not see a quality change. We additionally compare each
rater's direction judgment against the automatic DA metric on the same clip
pair; the two agree on $74.2\%$ of pairs with Cohen's $\kappa{=}0.58$, against
an inter-rater agreement of $0.71$, providing complementary agreement evidence; metric--human and inter-rater agreement are distinct comparisons.

\clearpage

\clearpage
\section{Failure Cases}
\label{sec:supp_failures}

\begin{figure}[!htbp]\centering
  \includegraphics[width=0.86\linewidth]{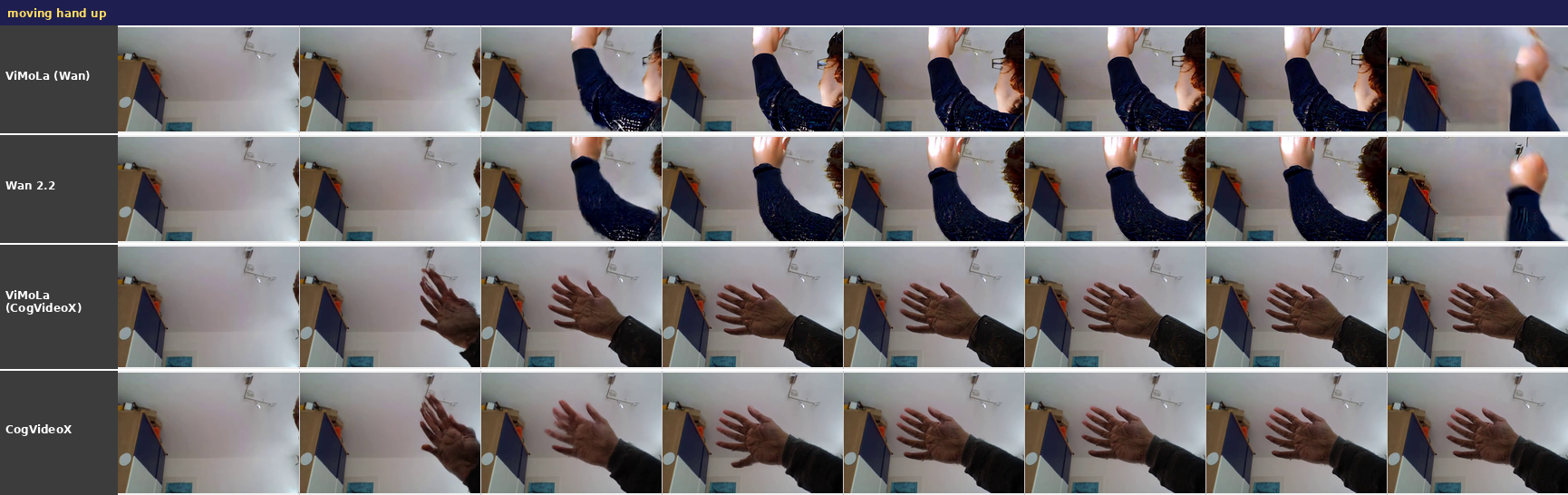}
  \caption{``moving hand up''}
\end{figure}

\begin{figure}[!htbp]\centering
  \includegraphics[width=0.86\linewidth]{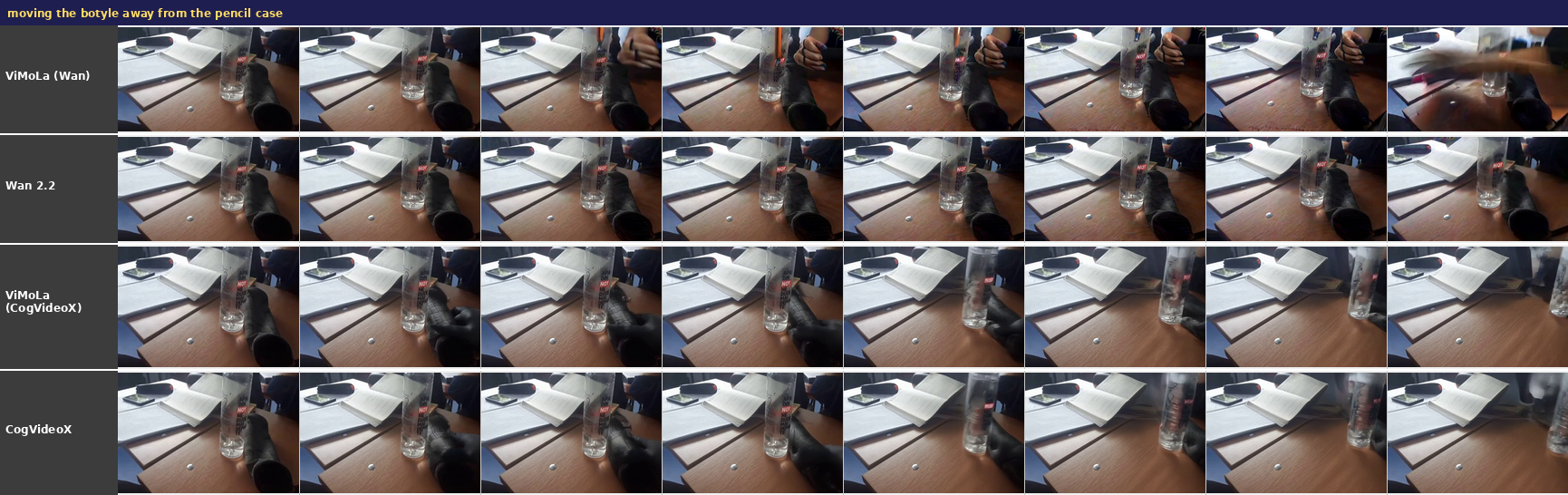}
  \caption{``moving the bottle away from the pencil case''}
\end{figure}

\begin{figure}[!htbp]\centering
  \includegraphics[width=0.86\linewidth]{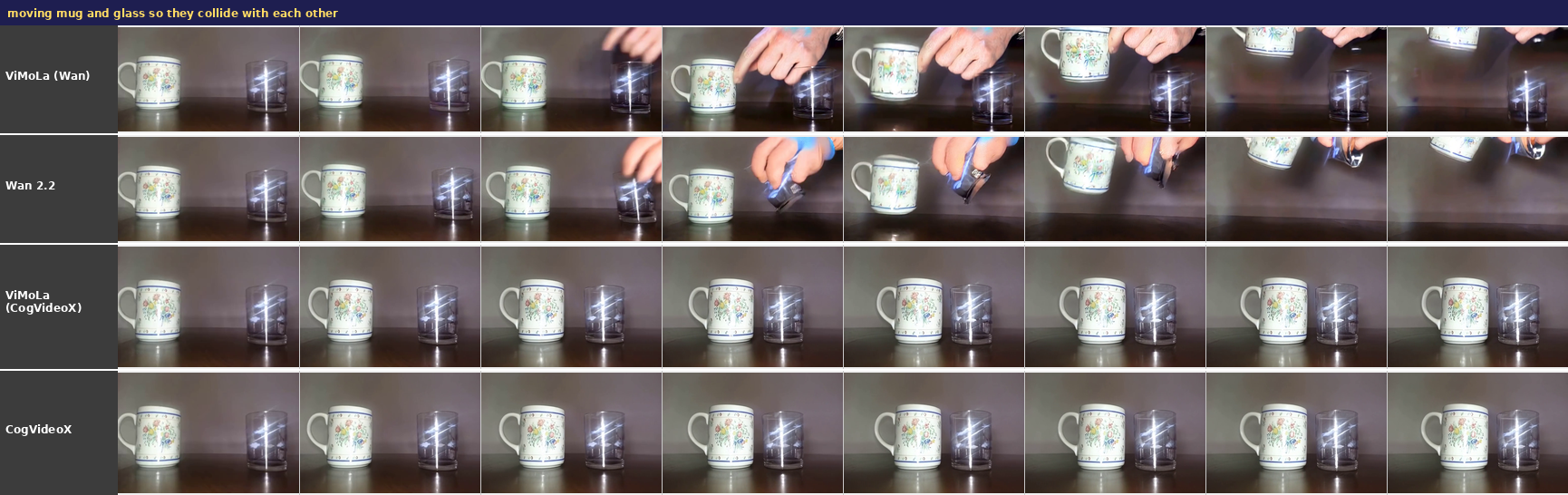}
  \caption{``moving mug and glass so they collide with each other''}
\end{figure}

The failures we observe cluster into three groups, all of which the quantitative
results anticipate. Compound prompts naming two successive motions are usually
realized as the first motion only, which is the behavior a single code per patch
predicts and which \cref{tab:complex} measures as the drop from $+0.187$ to
$+0.054$. Free-form prompts that do not name a direction produce a plausible but
arbitrary one, since the text-to-motion stage represents the ambiguity as a
distribution over codes but must commit to a single code at inference. Finally,
on clips whose foreground occupies a large fraction of the frame, the
foreground and background codes can conflict, producing a global drift; this may contribute to the subject-consistency gap of $0.731$ against $0.743$ in
\cref{tab:quality}, which is small on average but concentrated in these
cases.
\end{document}